\documentclass[runningheads,a4paper]{llncs}

\usepackage{geometry}
\usepackage{pgfplots}
\usepackage[all]{nowidow}
\usepackage[utf8]{inputenc}
\usepackage{tikz}
\usetikzlibrary{er,positioning, bayesnet}
\usepackage{multicol}
\usepackage{algpseudocode,algorithm,algorithmicx}

\usepackage{amsmath,amsfonts,bm}

\def\eqref#1{equation~\ref{#1}}

\def\eps{{\epsilon}}

\DeclareMathAlphabet{\mathsfit}{\encodingdefault}{\sfdefault}{m}{sl}
\SetMathAlphabet{\mathsfit}{bold}{\encodingdefault}{\sfdefault}{bx}{n}

\usepackage{shortcuts}

\usepackage{microtype}
\usepackage{graphicx}
\usepackage{subfigure}
\usepackage{booktabs}

\newtheorem{assumption}{Assumption}{\itshape}

\usepackage[textsize=tiny]{todonotes}
\usepackage{dsfont}
\usepackage[T1]{fontenc}

\usepackage{amsmath}
\usepackage{amssymb}
\usepackage{mathtools}

\usepackage{natbib}

\usepackage{amsfonts}
\usepackage{graphicx}
\usepackage{xr-hyper}
\usepackage{dsfont}
\usepackage{empheq}
\usepackage{enumitem}
\usepackage{bm}
\usepackage{bbm}
\usepackage{hhline}
\usepackage{stmaryrd}
\usepackage{tikz}      
\usetikzlibrary{calc, shapes, backgrounds}

\usepackage[utf8]{inputenc} 
\usepackage[T1]{fontenc}    
\usepackage{url}            
\usepackage{booktabs}       
\usepackage{amsfonts}       
\usepackage{nicefrac}       
\usepackage{epsfig}
\usepackage{color}
\usepackage{titletoc, setspace,
comment, relsize, float}
\usepackage{caption}
\usepackage{multirow}

\begin{document}

\title{Flexible Parametric Inference for \\Space-Time Hawkes Processes}


\author{Emilia Siviero\inst{1} \and
Guillaume Staerman\inst{2} \and
Stephan Clémençon\inst{1} \and Thomas Moreau\inst{2}
}
\authorrunning{Emilia Siviero}

%
%
\institute{LTCI, Télécom Paris, Institut Polytechnique de Paris, Palaiseau, 91120, France\\
\email{emilia.siviero@telecom-paris.fr}, \email{stephan.clemencon@telecom-paris.fr}\\ \and
Parietal Team, Université Paris-Saclay, Inria, CEA, Palaiseau, 91120, France\\
\email{guillaume.staerman@inria.fr}, \email{thomas.moreau@inria.fr}\\
}
\maketitle              
\thispagestyle{plain}\pagestyle{plain}


\begin{abstract}
    Many modern spatio-temporal data sets, in sociology, epidemiology or seismology, for example, exhibit self-exciting characteristics, triggering and clustering behaviors both at the same time, that a suitable Hawkes space-time process can accurately capture. This paper aims to develop a fast and flexible parametric inference technique to recover the parameters of the kernel functions involved in the intensity function of a space-time Hawkes process based on such data. Our statistical approach combines three key ingredients: 1) kernels with finite support are considered, 2) the space-time domain is appropriately discretized, and 3) (approximate) precomputations are used. The inference technique we propose then consists of a $\ell_2$ gradient-based solver that is fast and statistically accurate. In addition to describing the algorithmic aspects, numerical experiments have been carried out on synthetic and real spatio-temporal data, providing solid empirical evidence of the relevance of the proposed methodology.
\end{abstract}

\section{Introduction}\label{sec:intro}
Thanks to recent major technological advances in data collection and management (\textit{e.g.} the ubiquitous deployment of IoT or the widespread use of satellite imagery), spatio-temporal data is becoming increasingly available, see e.g. \cite{book_spatial}. Numerous applications could now be based on such observations, in a wide variety of fields ranging from sociology to environmental sciences through epidemiology for example, opening up new prospects for flexible space-time models, provided that effective methods for fitting them to the available data are developed of course. In particular, massive measurements sampled at an ever finer time and space granularity offer the possibility of observing complex (clustering and triggering) behaviors. Among point processes \cite{daley2003introduction}, Hawkes models \cite{Hawkes} have recently received a great deal of attention, as they take into account the self-exciting nature of observed events, space-time interaction and spatial anisotropy in a very flexible way. For well-chosen intensity functions, the probability of future events occurring over a given period and in a specific spatial proximity to past data increases with these point processes, see \cite{reinhart2018review}. As with the modeling of purely temporal phenomena with self-excitation characteristics \cite{laub2015hawkes, lewis2011nonparametric, ogata1988statistical, bompaire:tel-02316143}, these properties of space-time Hawkes processes are the main source of their interest in many fields requiring spatio-temporal analysis, such as epidemiology \cite{holbrook2022viral, kresin2022comparison, rambhatla2022toward, dong2023non}, criminology \cite{mohler2011self, mohler2014marked, dangelo, zhu2022spatiotemporal} or earthquake seismology \cite{ogata1998space, musmeci1992space, kwon2023flexible} for instance.  The main methodological challenge is then to design efficient inference techniques to fit Hawkes models to spatio-temporal datasets. To date, the literature has focused mainly on the estimation of intensity functions defined by separated space-time kernels often involving non-parametric or parametrized exponential kernels, see \cite{yuan2021fast}. However, non-parametric kernel estimation inevitably encounters limitations in terms of statistical efficiency, which are all the more important given the limited data available. Furthermore, exponential kernels are limited to scenarios where events do not instantaneously trigger other events.

\looseness=-1
Here, we develop a new methodology for general parametric inference for Space-Time Hawkes Processes (STHP).
Our algorithm is based on components similar to the ones used by \cite{staerman2023fadin} in the context of temporal events, extending them for spatio-temporal events.
It solves the modelization and numerical challenges posed by parametric STHP's complexity, enabling accounting for space-time interactions/inhomogeneity.
The method consists of a fast $\ell_2$ gradient-based solver, which provides flexibility in the choice of kernels, and is made computationally efficient based on the following three ideas.
First, we model the intensity function with bounded support parametric kernels.
We show that this choice offers significant advantages regarding computational complexity, allowing the use of discrete convolution and fast Fourier transform.
Second, the method leverages a discretization of the space-time domain, which can be seen as a hyperparameter to tune, whose selection depends on the underlying sampling precision of the data and data availability.
Combined with the first key component, it allows any choice of temporal and spatial kernels, providing flexibility in the modelization.
Third, our approach resorts to extensive use of precomputations, allowing for efficient implementation of the gradient-based inference procedure, with optimization steps independent of the number of events.
In this work, we show that the discretization has a low impact on the statistical performance of our estimator, and demonstrate the flexibility and efficiency of the proposed method on simulated and real-world datasets.

The paper is organized as follows. In \ref{sec:spatio_temp_hawkes}, we begin by reviewing briefly the main concepts relating to space-time Hawkes processes. Next, we detail the approach we propose to infer such models, discuss its numerical advantages and provide theoretical guarantees on the bias induced by the used discretization in \ref{sec:spatio_temp_fadin} . In \ref{sec:num}, we investigate the performance of the methodology promoted on simulated data from an empirical point of view. Finally, we present the experimental results obtained using our method based on earthquake real data in \ref{sec:app} for illustration purposes.

\section{Space-time Hawkes processes - background and framework}\label{sec:spatio_temp_hawkes}

We here briefly recall the notion of the STHP and its essential properties for our work.
We refer the reader to \cite{reinhart2018review} for a detailed account of these processes.
STHP are popular stochastic point processes that model the occurrence of events in both time and space with self-excitation.
Let $T \geq 0$ be a stopping time and $[0, T]$ the associated temporal observation period.
Let $\mathcal{S}\subset \mathbb{R}^2$ be a compact set of the space domain containing the locations of the observed events until time $T$.
A STHP realization consists of a set of distinct events: $\mathcal{H}_T = \{u_n=(x_n,y_n,t_n), (x_n,y_n)\in \mathcal{S},  t_n \in [0, T]\}$ occurring in continuous space-time, with an associated time $t_n$ and a location $(x_n, y_n)$.
The process's behavior is fully characterized by its \textit{intensity function}, which relies on the time and location of past events.
It can also include several types of events and is then called multivariate STHP. Given $D\geq 1$ type of events, for each $i\in \intervalleEntier{1}{D}$, we denote by $\mathbf{N}_i$ the random counting measure defined on $\mathcal{S} \times[0, T]$, such that $\mathbf{N}_i(\mathrm{d}x, \mathrm{d}y, \mathrm{d}t) = \sum\limits_{n=1}^{\infty} \delta_{(x_n, y_n, t_n)} (\mathrm{d}x, \mathrm{d}y, \mathrm{d}t)$, where $(x, y)$ and $t$ respectively denote the location and time of the events. The conditional intensity function is the instantaneous rate at which events are expected to occur. It is defined as the map from $\mathbb{R}^3$ to $\mathbb{R}_+$ such that: $\forall t \in [0, T], (x,y)\in \mathcal{S}$,
\begin{equation*}
    \lambda_i(x, y, t | \mathcal{H}_t) = \lim\limits_{\mathrm{d}x, \mathrm{d}y,\mathrm{d}t  \to 0} \frac{\mathbb{E} \big[ \mathbf{N}_i([x, x+\mathrm{d}x]\times [y, y+\mathrm{d}y]\times [t, t+\mathrm{d}t])| \mathcal{H}_t \big ]}{\mathrm{d}x \mathrm{d}y\mathrm{d}t }.
\end{equation*}

For a linear multivariate STHP, the intensity function of the $i$-th process has the following form:
\begin{align}\label{eq:cond_int}
    \lambda_i(x,y,t | \mathcal{H}_t) & = \boldsymbol{\mu}_i + \sum_{j=1}^{D}  \sum\limits_{u_n^j \in \mathcal{H}_t^j} \boldsymbol{\alpha}_{ij} \, g_{ij}(x-x_n^j, y - y_n^j, t-t_n^j),
\end{align}%
\looseness=-1
where the strictly positive $\boldsymbol{\mu}_i$'s are the {\it baseline} parameters, controling the spontaneous event apparition rate,
the positive $\boldsymbol{\alpha}_{ij}$ are the {\it excitation scaling} parameters,
and the $g_{ij}: \mathcal{S} \times [0, T] \mapsto \mathbb{R}_+$ are the spatio-temporal {\it kernel}, also referred as excitation functions.
The parameters $\boldsymbol{\alpha}_{ij}$ and $g_{ij}$ describe the excitation behavior between events.
Here we assume that $0<\boldsymbol{\alpha}_{ij}<1$ and $\int_{0}^{T}\int_{\mathcal{S}} g_{ij}(x,y,t) \mathrm{d}x\mathrm{d}y\mathrm{d}t = 1$, ensuring the stability of the generated process.
The statistical inference concerns the parameters $\boldsymbol{\mu}_i$, $\boldsymbol{\alpha}_{ij}$ and the triggering kernels that can be parametric \cite{yuan2019multivariate,reinhart2018review} or non-parametric \cite{lewis2011nonparametric,choi1999nonparametric,diggle1995second,kwon2023flexible}.
It is usually performed by minimizing the negative log-likelihood involving the intensity function.
In the literature \cite{mohler2014marked, yuan2019multivariate, ilhan2020modeling}, the kernel $g_{ij}(\cdot)$ is generally supposed to be first-order space-time separable (see \citep[Section 4.2]{gonzalez2016spatio} and \citep[Section 2.2]{reinhart2018review}), which means that the kernel is a product of time and spatial influences.
In addition, for computational purposes, the temporal kernel is often chosen as exponential, while the spatial kernel is chosen as Gaussian~\citep{mohler2014marked, yuan2019multivariate, ilhan2020modeling}.
These specific choices -- separability and constrained kernels-- both limit the modeling flexibility between the temporality and spatiality of events, reducing the applicability of STHP models to real-world data.

\section{A fast method for parametric inference in space-time Hawkes models}\label{sec:spatio_temp_fadin}

In this section, we outline the statistical approach promoted in this paper.
Building upon recent work by \cite{staerman2023fadin}, we develop a parametric inference framework that allows the estimation of any spatio-temporal kernel to reflect the underlying excitation of the process.
In particular, our approach shows linear scalability as a function of the number of events, greatly improving current methodologies.
Our inference procedure relies on three fundamental principles: (i) discretization, (ii) finite-support kernels for both spatial and temporal components of our processes and (iii) precomputation terms.
Moreover, in contrast to existing spatio-temporal literature, we leverage the overlooked ERM-inspired $\ell_2$ loss, which we extend to the spatial domain.
Although this loss benefits from advantageous precomputation terms, we propose an approximation for computationally intensive terms to accelerate the inference.
Finally, we provide theoretical guarantees concerning the influence of discretization on parameter estimation in~\ref{sec:discretization}.

\subsection{The inference approach - Key components}
\label{sec:key}

We now describe the main elements of our framework.
For simplicity, we assume that the spatial domain is a rectangle, \emph{i.e.}, of the form  $\mathcal{S}=\mathcal{X}\times \mathcal{Y}$ with $\mathcal{X}=[-S_\mathcal{X}, S_\mathcal{X}]$ and $\mathcal{Y}=[-S_\mathcal{Y}, S_\mathcal{Y}]$.
Our approach does not require the upper and lower limits to be identical or symmetrical with respect to zero for it to work.


\paragraph{Convolutional writing.}
For all $u=((x,y),t) \in \mathcal{S} \times [0, T]$ and for all $s \in [0, T]$, let $z^i_s(u) = \sum\limits_{u_n^i \in \mathcal{H}_s^i} \delta_{u_n^i}(u)$ be the sum of Dirac functions of event occurrences $u_n^i$, such that $z_s^i(u) = 1$ if $u \in \mathcal{H}_s^i$ and $0$ otherwise.
The intensity function \ref{eq:cond_int} can be reformulated as a convolution between the kernel $g_{ij}$ and $z_t^j$:
\begin{equation}\label{eq:conv}
    \forall u \in \mathcal{S} \times [0, T], \quad \, \lambda_i(u) = \boldsymbol{\mu}_i + \sum\limits_{j=1}^D \boldsymbol{\alpha}_{ij} \, (g_{ij} * z_t^j)(u).
\end{equation}

\paragraph{Finite-support kernels.} We consider the spatio-temporal kernels to be of finite length.  Let $W_{\mathcal{X}},W_{\mathcal{Y}}$ and $W_{\mathcal{T}}$ be the length of spatial and temporal supports. We assume that $\forall (x,y,t) \notin  [-W_{\mathcal{X}}, W_{\mathcal{X}}]\times [-W_{\mathcal{Y}}, W_{\mathcal{Y}}] \times [0, W_{\mathcal{T}}], \, g_{ij}(x, y, t) = 0$. Thus, any event $u_n^i=(x_n^i, y_n^i, t_n^i)$ may induce a new event only in
\begin{equation*}
    [x_n^i-W_{\mathcal{X}},x_n^i+ W_{\mathcal{X}}] \times [y_n^i -W_{\mathcal{Y}},y_n^i+ W_{\mathcal{Y}}] \times [t_n^i, t_n^i+W_{\mathcal{T}}]\enspace.
\end{equation*}

In combination with \ref{eq:conv}, the main advantage of this assumption is to make it possible to leverage discrete convolution and fast Fourier transform for efficient intensity computation.

\paragraph{Discretization.} Discretization has been successfully used in Hawkes processes \cite{kirchner2017estimation} and recently in spatio-temporal processes \cite{sheen2022tensor}. Given the spatial support as a compact set $[-S_\mathcal{X}, S_\mathcal{X}]\times  [-S_\mathcal{Y}, S_\mathcal{Y}]\subset\mathbb{R}^2$, we propose to define a three dimensional regular grid $\mathcal{G}=\mathcal{G}_{\mathcal{X}} \times \mathcal{G}_{\mathcal{Y}} \times \mathcal{G}_T $ such that $\mathcal{G}_{\mathcal{X}} = \{-S_\mathcal{X}, -S_\mathcal{X}+\Delta_{\mathcal{X}}, \cdots, -S_\mathcal{X} + G_{\mathcal{X}} \Delta_{\mathcal{X}}\}$, $\mathcal{G}_{\mathcal{Y}} = \{-S_\mathcal{Y},-S_\mathcal{Y} +  \Delta_{\mathcal{Y}}, \cdots, -S_\mathcal{Y} + G_{\mathcal{Y}} \Delta_{\mathcal{Y}} \}$ and  $\mathcal{G}_T = \{0, \Delta_T, \cdots, G_T \Delta_T\}$ with  $G_T \Delta_T = T$, $G_{\mathcal{X}} \Delta_{\mathcal{X}} = 2 S_{\mathcal{X}}$, $G_{\mathcal{Y}} \Delta_{\mathcal{Y}} = 2 S_{\mathcal{Y}}$,  and $\Delta_{\mathcal{X}}, \Delta_{\mathcal{Y}}, \Delta_T > 0$ are the stepsizes of the spatial and  temporal grids.  Further, we project the observed events on these grids and define
 $\widetilde{\mathcal{H}}_{T}^i$ as the projected space-time stamps of $\mathcal{H}_{T}^i$.  Given $ v = (v_x, v_y, v_t) \in  \llbracket 0, G_{\mathcal{X}} \rrbracket \times  \llbracket 0, G_{\mathcal{Y}} \rrbracket \times \llbracket 0,  G_T \rrbracket$, we define the vector versions $g_{ij}^{\Delta}[v] = g_{ij}(v \Delta)$ of the kernels, and the sparse vector of events:

 {\small
 $$z_{t}^j[v]  = \#\{(x^j_n, y^j_n, t^j_n): \;  |x^j_n - (-S_{\mathcal{X}}+ v_x\Delta_\mathcal{X})| \leq \frac{\Delta_\mathcal{X}}{2}, \, |y^j_n - (-S_{\mathcal{Y}} + v_y\Delta_\mathcal{Y})| \leq \frac{\Delta_\mathcal{Y}}{2}, \, |t^j_n - v_t\Delta_T| \leq \frac{\Delta_T}{2}\},$$ }

 that reflects the number of events projected at the position $v$ on the grid $\mathcal{G}$. With these notations, we can rewrite the intensity function of the $i^{th}$ process of our discretized STHP  relying on discrete convolution such that for any $ v = (v_x, v_y, v_t) \in   \llbracket 0, G_{\mathcal{X}} \rrbracket \times  \llbracket 0, G_{\mathcal{Y}} \rrbracket \times \llbracket 0,  G_T \rrbracket$, we have:

\begin{align}
    \Tilde{\lambda}_i[v] & = \boldsymbol{\mu}_i + \sum\limits_{j=1}^D \boldsymbol{\alpha}_{ij} \, (g_{ij}^{\Delta} * z_{v_t \Delta_T}^j)[v] \nonumber  = \boldsymbol{\mu}_i + \sum\limits_{j=1}^D \sum\limits_{\substack{\tau_x=1}}^{L_{\mathcal{X}}}\sum\limits_{\substack{\tau_y=1}}^{L_{\mathcal{Y}}} \sum\limits_{\tau_t=1}^{L_T} \boldsymbol{\alpha}_{ij} \, g_{ij}^{\Delta}\left[\tau\right] z_{v_t \Delta_T}^j\left[v - \kappa\right],
\end{align}

where  $\tau = (\tau_x, \tau_y, \tau_t)$, $\kappa = \left(\tau_x - l_{\mathcal{X}}, \tau_y - l_{\mathcal{Y}}, \tau_t\right)$, with $l_{\mathcal{X}} = \left\lfloor L_{\mathcal{X}} /2 \right\rfloor + 1$, $l_{\mathcal{Y}} = \left\lfloor L_{\mathcal{Y}} /2\right\rfloor + 1$. Define $L_T = \left\lfloor W_T / \Delta_T\right\rfloor + 1$ the number of points on the discretized temporal support, and $L_{\mathcal{X}} =  \left\lfloor 2 W_{\mathcal{X}} /\Delta_{\mathcal{X}}\right\rfloor + 1$, $L_{\mathcal{Y}} =  \left\lfloor 2 W_{\mathcal{Y}} / \Delta_{\mathcal{Y}}\right\rfloor + 1$ the number of points on each component of the discretized spatial support.

\subsection{Efficient inference with Empirical Risk Minimization}\label{subsec:comp}

 While the spatio-temporal literature focuses on the negative log-likelihood minimization to infer Hawkes parameters, we decide to focus on the ERM-inspired least squares loss \citep{reynaud2010near,Reynaud-Bouret2014,bacry2020sparse} only used in classical temporal Hawkes process so far. In contrast to the log-likelihood, the least square loss disentangles the computation dependency in the number of events from the optimization procedure. Indeed, it involves precomputation terms, that will summarize the offset information of events, thanks to the absence of logarithm in the right part of the loss. Let  $\mathcal{H}_{T} = \{\mathcal{H}_{T}^i\}_{i=1}^D$ be a set of observed spatio-temporal events. Assuming a class of spatio-temporal parametric kernels parametrized by $\boldsymbol{\eta}_{ij}$, the objective is to find $\boldsymbol{\theta}=\{\boldsymbol{\mu}_i, \boldsymbol{\alpha}_{ij},\boldsymbol{\eta}_{ij}\}_{i,j}$  that minimizes:
\begin{align}
    &\mathcal{L}(\boldsymbol{\theta}, \mathcal{H}_{T}) =  \sum\limits_{i=1}^D \left( \int_0^T \int_\mathcal{S} \lambda_i(x, y, t)^2~ \mathrm{d}x \mathrm{d}y \mathrm{d}t - 2 \sum\limits_{u_n^i \in \mathcal{H}_{ T}^i} \lambda_i(x_n^i, y_n^i, t_n^i)\right).
    \label{eq:loss}
\end{align}

Given the core components of our model described in ~\ref{sec:key}, the objective is then to minimize the discretized $\ell_2$ loss defined by:

\begin{align*}
    &\mathcal{L}_{\mathcal{G}}(\boldsymbol{\theta}, \widetilde{\mathcal{H}}_{T}) =  \sum\limits_{i=1}^D \left(\Delta_{\mathcal{X}}\Delta_{\mathcal{Y}} \Delta_T \sum\limits_{v_x = 0}^{G_{\mathcal{X}}}  \sum\limits_{v_y = 0}^{G_{\mathcal{Y}}}\sum\limits_{v_t = 0}^{G_T} \left(\Tilde{\lambda}_i[v_x, v_y, v_t]\right)^2 - 2 \sum\limits_{\Tilde{u}_n^i \in \widetilde{\mathcal{H}}_{ T}^i} \Tilde{\lambda}_i\left[\frac{\Tilde{x}_n^i}{\Delta_{\mathcal{X}}}, \frac{\Tilde{y}_n^i}{\Delta_{\mathcal{Y}}}, \frac{\Tilde{t}_n^i}{\Delta_T} \right]\right).
\end{align*}
This approximates the integral in \ref{eq:loss} by a sum on the grid $\mathcal{G}$  after projecting the space-time stamps of $\mathcal{H}_T$ on it.
By developing and rearranging the terms in the discretized loss above, one can see some constants that do not depend on $\boldsymbol{\theta}$ and thus can be precomputed:
\begin{align*}
     \mathcal{L}_{\mathcal{G}}(\boldsymbol{\theta}, \widetilde{\mathcal{H}}_{T}) = &(T+\Delta_T) (2 S_{\mathcal{X}} + \Delta_{\mathcal{X}}) (2 S_{\mathcal{Y}} + \Delta_{\mathcal{Y}})\sum\limits_{i=1}^D \boldsymbol{\mu}_i^2 \\&
     + 2 \Delta_{\mathcal{X}} \Delta_{\mathcal{Y}}\Delta_T \sum\limits_{i=1}^D \boldsymbol{\mu}_i \sum\limits_{j=1}^D \sum\limits_{\tau_x=1}^{L_{\mathcal{X}}} \sum\limits_{ \tau_y=1}^{L_{\mathcal{Y}}} \sum\limits_{\tau_t=1}^{L_T} \boldsymbol{\alpha}_{ij} \, g_{ij}^{\Delta}\left[\tau\right] \Phi_j(\tau; G)\\ &
    + \Delta_{\mathcal{X}} \Delta_{\mathcal{Y}} \Delta_T \sum\limits_{i, j, k =1}^D \sum\limits_{\tau_x, \tau_x' = 1}^{L_{\mathcal{X}}} \sum\limits_{\tau_y, \tau_y' = 1}^{L_{\mathcal{Y}}} \sum\limits_{\tau_t, \tau_t'=1}^{L_T} \boldsymbol{\alpha}_{ij} \, \boldsymbol{\alpha}_{ik} \, g_{ij}^{\Delta}\left[\tau\right] g_{ik}^{\Delta}\left[\tau'\right] \Psi_{j, k}(\tau, \tau'; G)\\ &
    - 2 \sum\limits_{i=1}^D \left(N_T^i \boldsymbol{\mu}_i + \sum\limits_{j=1}^D\sum\limits_{\tau_x=1}^{L_{\mathcal{X}}} \sum\limits_{ \tau_y=1}^{L_{\mathcal{Y}}}  \sum\limits_{\tau_t=1}^{L_T} \boldsymbol{\alpha}_{ij} \, g_{ij}^{\Delta}\left[\tau\right] \Phi_j(\tau; \widetilde{\mathcal{H}}_{T}^i)\right),
\end{align*}
where the following terms can be precomputed:

\begin{itemize}
    \item $\Phi_j(\tau; G) = \sum\limits_{v_x=0}^{G_{\mathcal{X}}}\sum\limits_{v_y=0}^{G_{\mathcal{Y}}}  \sum\limits_{v_t = 0}^{G_T} z_{v_t \Delta_T}^j\left[v - \kappa\right]$, \qquad \qquad \tiny{$\bullet$} \normalsize{\, $\Phi_j(\tau; \widetilde{\mathcal{H}}_{T}^i) = \sum\limits_{\Tilde{u}_n^i \in \Tilde{\mathcal{H}}_{T}^i} z_{\Tilde{t}_n^i}^j\left[\frac{\Tilde{u}_n^i}{\Delta} - \kappa\right]$,}
    \item $\Psi_{j, k}(\tau, \tau'; G) = \sum\limits_{v_x=0}^{G_{\mathcal{X}}}\sum\limits_{v_y=0}^{G_{\mathcal{Y}}}  \sum\limits_{v_t = 0}^{G_T} z_{v_t \Delta_T}^j\left[v -\kappa\right] z_{v_t \Delta_T}^{k}\left[v - \kappa'\right]$,
\end{itemize}

with $\kappa = \left(\tau_x - l_{\mathcal{X}}, \tau_y - l_{\mathcal{Y}}, \tau_t\right)$ and $G = (G_{\mathcal{X}}, G_{\mathcal{X}}, G_T)$. $\Phi_j(\tau; G)$ defines the total number of events of the $j$-th process by removing a part of the grid of size  $\kappa$. $\Psi_{j, k}(\tau, \tau'; G)$ denotes how many events of the $j$-th process with a lag $\kappa$ are matching the events of the $k$-th process with a lag $\kappa'$. $\Phi_j(\tau; \widetilde{\mathcal{H}}_{T}^i)$ assess how many events in the $j$-th process are at the same position than events of the $i$-th process with a lag $\kappa$. As these three terms do not depend on the set of parameters, they can be precomputed at initialization and used at each step of the optimization procedure. Let $\overline{G}=G_{\mathcal{X}}G_{\mathcal{Y}}G_T$ be the total number of element on the grid $\mathcal{G}$ and $\overline{L}=L_{\mathcal{X}}L_{\mathcal{Y}}L_T$ be the total number of discretization points of the kernels $g_{ij}$. The term $\Psi_{j, k}(\tau, \tau'; G) $ is the bottleneck of these precomputations and requires $O(\overline{G})$ for each tuples $(\tau, \tau')$ and $(j,k)$. Thus, it leads to a total computational complexity of $O(D^2\overline{L}^2 \overline{G})$. This may be limiting in the choices of the discretization steps $\Delta_{\mathcal{X}},\Delta_{\mathcal{Y}}$ and $\Delta_T$, driving the user to take them not too small and then inducing discretization bias in the results of the solver.

\paragraph{Approximation of $\Psi$.}
The precomputation terms are computed only once, but in the spatio-temporal setting, they may suffer from a computational burden.
The bottleneck is the $\overline{L}^2$ presence in the computational complexity of $\Psi_{i,j}(\cdot~; \cdot)$. Here, we provide an approximation of $\Psi_{i,j}$, denoted by $\widetilde{\Psi}_{i,j}$, to alleviate this computation challenge. Precisely, we propose

\begin{align*}
    \widetilde{\Psi}_{j,k}(\tau; G) = \sum\limits_{v_x=0}^{G_{\mathcal{X}}}\sum\limits_{v_y=0}^{G_{\mathcal{Y}}}  \sum\limits_{v_t = 0}^{G_T} z_{v_t \Delta_T}^j\left[v_x, v_y, v_t \right] z_{v_t \Delta_T}^{k}\left[v_x -\tau_x, v_y-\tau_y, v_t - \tau_t \right].
\end{align*}

The $\widetilde{\Psi}_{j,k}(\tau; G)$ is the number of events in the discretized $j$-th process that have events with a lag $\tau$ in the discretized $k$-th process.
Thus, it will be evaluated in $\tau-\tau'$ in the third term of the loss $\mathcal{L}_{\mathcal{G}}$. The quadratic complexity in $\overline{L}$ is then removed and makes the computation of $\widetilde{\Psi}_{j,k}$ of order $O(\overline{L}\overline{G})$, which is linear with the grid discretization and the kernel grids and comparable to the computation complexity of $\Phi_j(\cdot; \cdot)$. The loss of information in this approximation lies in the boarding effects of the grid, which are small if the domain size is large in front of the kernel support.


\paragraph{Gradient-based optimization.} The inference procedure employs gradient descent to minimize the $\ell_2$ loss function $\mathcal{L}_{\mathcal{G}}$. Our approach design enables the utilization of flexible parametric kernels for both temporal and spatial patterns. It efficiently computes exact gradients for each kernel parameter, assuming the kernel is both differentiable and possesses finite support. Consequently, gradient-based optimization methods can be applied without constraints, in stark contrast to the EM algorithm, widely used in the literature, necessitating a closed-form solution to nullify the gradient—a challenge with numerous kernels. It's worth noting that this problem typically entails non-convexity, potentially leading to convergence towards local minima. The computations of the gradients are detailed in \ref{sec:grad}.

\subsection{On the bias of spatio-temporal discretization - theoretical guarantees} \label{sec:discretization}

Discretization introduces a perturbation in the loss value. Here, we assess the impact of this perturbation on parameter estimation as $\Delta_{\mathcal{X}},\Delta_{\mathcal{Y}}$ and $\Delta_T$ approach $0$. Throughout this section, we consider a set of events $\mathcal{H}_T$ stemming from a spatio-temporal Hawkes process with intensity functions expressed in the parametric form $\lambda_i(\cdot; \boldsymbol{\theta}^*)$, with $\boldsymbol{\theta}^*=\{\boldsymbol{\mu}_i^*, \boldsymbol{\alpha}_{ij}^*, \boldsymbol{\eta}_{ij}^* \}_{i,j}$. It is important to note that if the intensity of the process $\mathcal{H}_T$ does not belong to the parametric family $\lambda_i(\cdot; \boldsymbol{\theta})$, then $\boldsymbol{\theta}^*$ is defined as the best approximation of its intensity function in the $\ell_2$ sense. The objective of the inference process is to estimate the parameters $\boldsymbol{\theta}^*$.

When working with the projected set of events $\widetilde{\mathcal{H}}_T$, the original tuple $(x_n^i, y_n^i, t_n^i)$ is replaced with its projection on the grid $\mathcal{G}$, denoted as $\tilde{x}_n^i = x_n^i + \delta_{x, n}^i, \tilde{y}_n^i = y_n^i + \delta_{y, n}^i, \tilde{t}_n^i = t_n^i + \delta_{t, n}^i$. Here, $\delta_{x, n}^i$ is uniformly distributed over the interval $[-\Delta_{\mathcal{X}}/2, \Delta_{\mathcal{X}}/2]$, $ \delta_{y, n}^i$ over $[-\Delta_{\mathcal{Y}}/2, \Delta_{\mathcal{Y}}/2]$ and $\delta_{t, n}^i$ over $[-\Delta_T/2, \Delta_T/2]$. We define the discrete estimator $\widehat{\boldsymbol{\theta}}_\Delta$ as $\widehat{\boldsymbol{\theta}}_\Delta = \arg\min_{\boldsymbol{\theta}} \mathcal{L}_{\mathcal{G}}(\boldsymbol{\theta}, \widetilde{\mathcal{H}}_T)$, the set of parameters minimizing the discrete loss. The error induced by $\widehat{\boldsymbol{\theta}}_\Delta$ can be upper-bounded as follows:

\begin{equation}
\|\widehat{\boldsymbol{\theta}}_{\Delta} - \boldsymbol{\theta}^* \|_2
\le \underbrace{\|\widehat{\boldsymbol{\theta}}_c - \boldsymbol{\theta}^*\|_2}_{(1)}
+ \underbrace{\|\widehat{\boldsymbol{\theta}}_{\Delta} - \widehat{\boldsymbol{\theta}}_c\|_2}_{(2)},
\end{equation}
where $\widehat{\boldsymbol{\theta}}_c = \arg\min_{\boldsymbol{\theta}} ~\mathcal{L}\pars{\boldsymbol{\theta}, \mathcal{H}_T}$  is the reference estimator for $\boldsymbol{\theta}^*$ based on the standard $\ell_2$ estimator for continuous spatio-temporal Hawkes processes. This decomposition involves the statistical error $(1)$ and the bias error induced by the discretization $(2)$. The statistical term $(1)$ measures the deviation of the parameters obtained by minimizing the $\ell_2$ continuous loss from the true parameters, given a finite amount of data. In contrast, the term $(2)$ represents the discretization bias induced by minimizing the discrete loss instead of the continuous one.

In the following proposition, we focus on the discretization error $(2)$, which relates to the computational trade-off offered by our method. Before stating our proposition, we need further assumption on the discretized grid, implying that no event collapses on the same grid element.

\begin{assumption}\label{ass:1}
Suppose  for any $i,j \in \intervalleEntier{1}{D}$, we have  $\Delta_{\mathcal{X}} < \min\limits_{x_n^i, x_m^j \in \mathcal H_T} |x_n^i -x_m^j|$,  $\Delta_{\mathcal{Y}} < \min\limits_{y_n^i, y_m^j \in \mathcal H_T} |y_n^i -y_m^j|$ and  $\Delta_T < \min\limits_{t_n^i, t_m^j \in \mathcal H_T} |t_n^i -t_m^j|$.
\end{assumption}

We now study the perturbation of the loss due to discretization.

\begin{proposition}
\label{prop:bias}
Let $\mathcal H_T$ and $\widetilde{\mathcal H}_T$ be respectively a set of events (drawn from a spatio-temporal Hawkes process) and its discretized version on the grid $\mathcal{G}$ with stepsize $\Delta=(\Delta_{\mathcal{X}}, \Delta_{\mathcal{Y}}, \Delta_T)$. Suppose \ref{ass:1} to be satisfied. Thus, for any  $v=(v_x, v_y, v_t)$,  it holds:

\begin{equation*}
        \widetilde \lambda_i[v] = \lambda_i(v\Delta) - \sum_{j=1}^p \sum_{u_m^{j} \in \mathcal{H}_{v\Delta}^j} \hspace{-0.2cm}\delta_m^j . \nabla_u g_{ij}(v\Delta - u_m^j)
    + O\pars{\|\Delta\|_2^2},
\end{equation*}

and
{\small
\begin{align*}
    \mathcal{L}(\boldsymbol{\theta}, \widetilde{\mathcal{H}}_{T}) \hspace{-0.05cm} \leq \hspace{-0.05cm} \mathcal{L}(\boldsymbol{\theta}, \mathcal{H}_{T}) \hspace{-0.05cm}+ \hspace{-0.05cm}\|\Delta\|_2\sum_{i=1}C(\lambda_i) \hspace{-0.05cm}+\hspace{-0.05cm} 2 \sum_{i,j}\hspace{-0.1cm}\sum_{ \substack{u_n^{i} \in \mathcal{H}_T^i \\ u_m^{j} \in \mathcal{H}_T^j} }\hspace{-0.2cm}\pars{\delta_m^j - \delta_n^i}
    . \nabla_u g_{ij}\pars{u_n^i - u_m^j} \hspace{-0.05cm}+\hspace{-0.05cm} O\pars{\|\Delta\|_2^2},
\end{align*}}
where $C(\lambda_i)$ is a constant depending only on the regularity of $\lambda_i$.
\end{proposition}

The technical proof is provided in \ref{subsec:prop:bias}. The first result follows directly from the Taylor expansion of the intensity for the kernels. For the loss, the initial perturbation term, $\|\Delta\|_2\sum_{i=1}C(\lambda_i)$, arises from approximating the integral with a finite Euler sum by the generalization of the Koksma-Hlawka inequality for piece-wise smooth functions \cite{brandolini2013koksma}, while the second term stems from the perturbation of the intensity. This proposition demonstrates that, as the norm of the discretization steps $\|\Delta\|$ approaches 0, the perturbed intensity and the $\ell_2$ loss serve as accurate estimates of their continuous counterparts. We now proceed to quantify the discretization error $(2)$ as $\|\Delta\|$ goes to 0.

\begin{proposition} \label{prop2}
Suppose the assumption in \ref{prop:bias} is satisfied.
Then, if the estimators $\widehat{\boldsymbol{\theta}}_c = \arg\min_{\boldsymbol{\theta}} \mathcal{L}(\boldsymbol{\theta}, \mathcal{H}_T)$ and $\widehat{\boldsymbol{\theta}}_\Delta = \arg\min_{\boldsymbol{\theta}} \mathcal{L}_{\mathcal{G}}(\boldsymbol{\theta}, \widetilde{\mathcal{H}}_T)$ are uniquely defined, $\widehat{\boldsymbol{\theta}}_\Delta$ converges to $\widehat{\boldsymbol{\theta}}_c$ as $\|\Delta\| \to 0$. Moreover, if $\mathcal L$ is $C^2$ and its hessian $\nabla^2\mathcal L\pars{\widehat{\boldsymbol{\theta}}_c}$ is positive definite with $\varepsilon > 0$ its smallest eigenvalue, then
$
    \norme{\widehat{\boldsymbol{\theta}}_{\Delta} - \widehat{\boldsymbol{\theta}}_c }_2 \le \dfrac{\mathrm{max} \{\|\Delta\|_2, \|\Delta\|_{\infty} \} }{\varepsilon}~\omega\pars{\widehat{\boldsymbol{\theta}}_\Delta}
$
, with $\omega\pars{\widehat{\boldsymbol{\theta}}_\Delta} = O(1)$.
\end{proposition}

The technical proof is provided in \ref{subsec:prop2}.
In contrast to the bound in \cite{staerman2023fadin} that does not consider the spatial components, the asymptotic rates we obtained depend on $\Delta_{\mathcal{X}}, \Delta_{\mathcal{Y}}$ and $\Delta_T$. It shows that they all must go towards zero to have $\boldsymbol{\theta}_\Delta$ converging to the continuous estimates. When $\Delta_{\mathcal{X}}, \Delta_{\mathcal{Y}}$ and $\Delta_T$ go to zero, the provided rate is linear w.r.t. the stepsize grid parameters, allowing fast convergence. However, it also shows that if one of the grids is not refined enough, it may deteriorate the performance of the discretized estimator.
Another important remark is that the rate only depends on the sum of the discretization stepsize, and does involve their product. This means that we can use discretization of the same order for each dimension without having incurring a large degradation of the statistical efficiency of the estimator.

\section{Numerical experiments}\label{sec:num}

In this section, we present several numerical experiments on synthetic data. First, we analyze the influence of the discretization bias on the accuracy of the model as well as the computation time of the proposed approach. Then, we assess its statistical accuracy with various values for the stopping time $T$ and the set of locations $\mathcal{S}$. Finally, we investigate the accuracy and the computation time of the developed precomputation approximation. The codes used for this study are publicly available at \url{https://github.com/EmiliaSiv/Flexible-Parametric-Inference-for-Space-Time-Hawkes-Processes}.

\subsection{On the bias of spatio-temporal discretization - numerical assessment}\label{subsec:num_exp_dis}

\begin{figure}[!ht]
    \centering
    \begin{tabular}{cc}
    \multicolumn{2}{c}{\hspace{0.6cm} \includegraphics[scale=0.3]{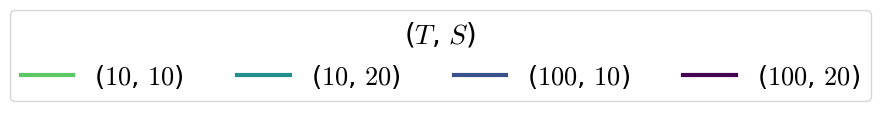}} \\
         \includegraphics[scale=0.4, trim=0cm 0 0 0]{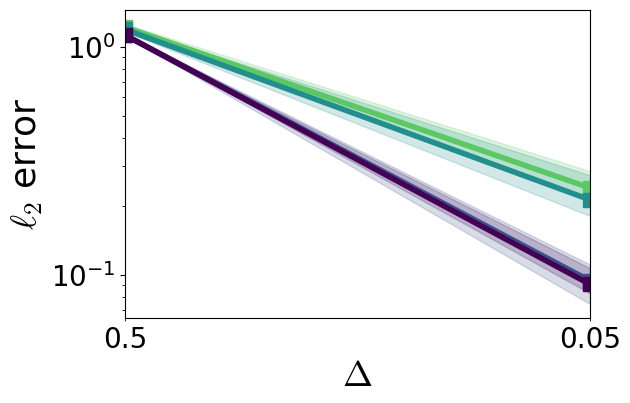} & \includegraphics[scale=0.4, trim=0cm 0 0 0]{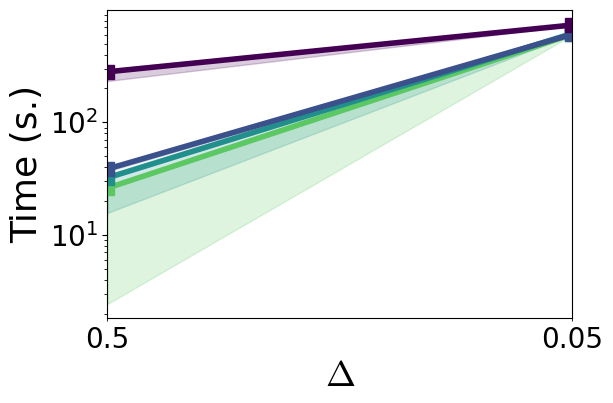}
           \end{tabular}
    \caption{Median and $25$\%-$75$\% quantiles of the $\ell_2$-norm between true and estimated parameters (left), and computational time with respect to $\Delta$ (right), for various  $T$ and $S$.}
    \label{fig:exp1_kur}
\end{figure}

To study the estimation error induced by the discretization, we run four experiments and report their results in \ref{fig:exp1_kur}, \ref{fig:exp1_Exp_Gauss}, and \ref{fig:exp1_POW_KUR}.
We consider a one-dimensional STHP with intensity function described as in \ref{eq:cond_int}. We simulate events according to the Immigration-Birth algorithm \citep{moller2005perfect} with a stopping time $T$ and a square spatial support with bound $S$. The excitation kernel $g$ is defined as a time-space separated kernel such that $g(x, y, t) = h(x, y)f(t)$. We set the baseline parameter $\mu=0.5$ and the scaling excitation factor $\alpha=0.6$. The spatial kernel is defined as a truncated Gaussian  with finite support $[-1,1]^2$ defined as

$$h(x, y; m, \sigma) \propto  \exp \left(-\frac{(x-m_1)^2 + (y-m_2)^2}{2 \sigma^2}\right) \mathbb{I} \{(x, y) \in [-1, 1]^2 \},$$

where the mean is $m=(m_1,m_2)=(0,0)$ and the  standard deviation is $\sigma=0.1$. For the temporal triggering function, we consider the Kumaraswamy density function, defined as

\begin{equation*}
    f(t; a, b) =  abt^{a-1}(1-t^a)^{b-1} \mathbb{I} \{0 \leq t \leq 1 \},
\end{equation*}

where $a=2$ and $b=2$. Hence, the set of parameters to estimate is $\boldsymbol{\theta}^* = (\mu, \alpha, m, \sigma, a, b)$.
We compute the estimates of $\boldsymbol{\theta}^*$ for varying stepsizes $\Delta = (\Delta_{\mathcal{X}}, \Delta_{\mathcal{Y}}, \Delta_T)$ of the spatial and temporal grids. To highlight \ref{prop2}, we set equal refinement of the grid w.r.t. each modality, i.e., $\Delta_T = \Delta_{\mathcal{X}} = \Delta_{\mathcal{Y}} \in [0.5, 0.05]$. The experiments are computed for multiple ending time $T\in \{10, 100 \}$ and spatial bounds $S\in \{ 10, 20 \}$. Our estimates $\hat{\boldsymbol{\theta}}$, obtained by applying our approach, are compared to $\boldsymbol{\theta}^*$. Precisely, the median (over $100$ runs) and the $25$\%-$75$\% quantiles of the $\ell_2$ estimation error $\| \hat{\boldsymbol{\theta}}- \boldsymbol{\theta}^* \|_2$ are displayed in \ref{fig:exp1_kur} (left). The associated computation time is depicted in \ref{fig:exp1_kur} (right).

One can observe that the estimation error goes towards zero as $\Delta$ decreases and supports the theoretical rates obtained in \ref{prop2}. In addition, when $T$ and $S$ increases, i.e. the number of events increases, the error diminishes. The computation time is efficient according to the setting size and grows as  $T$ and $S$  increase. As expected, the spatial bound adds more computation than the temporal one.

\subsection{On the statistical error}\label{subsec:num_exp_comp}

The statistical error of a STHP is challenging to assess theoretically. To that end, we investigate the statistical error returned by the parameters estimations of our approach based on the values of the ending time $T$ and of the spatial bounds $S$, assuming a square spatial support.

\begin{figure}[!ht]
    \centering
    \begin{tabular}{cc}

         \includegraphics[scale=0.4, trim=0cm 0 0 0]{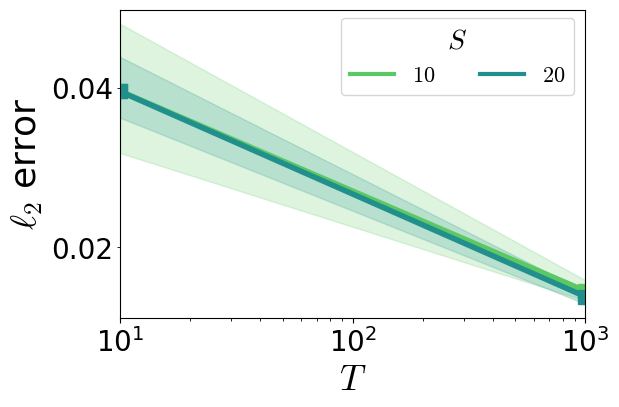} & \includegraphics[scale=0.4, trim=0cm 0 0 0]{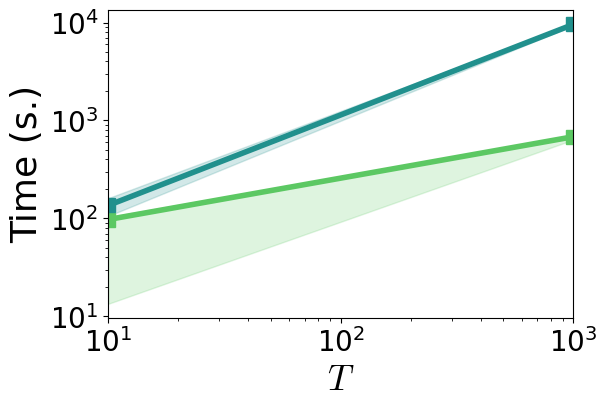}
           \end{tabular}
    \caption{Median and $25$\%-$75$\% quantiles of the $\ell_2$-norm between true and estimated parameters (left), and computational time with respect to $T$ (right), for various $S$.}
    \label{fig:exp2_time_T}
\end{figure}

We simulate a one dimensional STHP with a truncated Gaussian spatial kernel defined as in \ref{subsec:num_exp_dis}. For the temporal kernel, we set a truncated Gaussian defined as

$$f(t) \propto \exp\left(\frac{(t-m_T)^2}{2 \sigma_T^2}\right) \mathbb{I} \{0 \leq t \leq W_T \},$$
where the mean is $m_T = 0.5$, the standard deviation is $\sigma_T = 0.1$ and the finite support length is $W_T=1$. Events are simulated with varying end time and spatial bounds, i.e.,  $T \in [10, 1000]$ and $S \in \{10, 20\}$. We compute our proposed approach by fixing $\Delta = (0.1, 0.1, 0.1)$ since we are no longer interested in the discretization bias. We report the median (over $100$ runs) and the $25$\%-$75$\% quantiles of the $\ell_2$ estimation error $\| \hat{\boldsymbol{\theta}}- \boldsymbol{\theta}^* \|_2$ in \ref{fig:exp2_time_T} (left), alongside with the computation time with respect to $T$ and $S$ (right).

We observe that the $\ell_2$-norm goes towards zero as $T$ increases. We can see that the spatial bound value has influence on the variance of the error but the convergence is identical w.r.t. the median.

\subsection{Approximation of the bottleneck precomputation term}

The experiment in this part supports the choice of the approximation of $\Psi$ discussed in  \ref{subsec:comp}. We simulate data with the same setting than in \ref{subsec:num_exp_comp}. In order to show the relevance of the chosen approximation, we computed the true precomputation, denoted by $\Psi^*$, and the approximated one $\widetilde{\Psi}$ for various sizes of $T$ and $S$. Note that, due to the computational burden from $\Psi$, the values of $T$ and $S$ are small. Here, we assess the relative approximation error between $\Psi^*$ and $\widetilde{\Psi}$ with two metrics: the 1-norm $\|\cdot\|_1$ and the Frobenius norm $\|\cdot \|_F$ between tensors. The results are reported in \ref{tab:approx_psi} together with their computation time.

{\renewcommand{\arraystretch}{1.5} 
{\setlength{\tabcolsep}{0.7cm}
\begin{table*}[!h]
    \centering
    
    \caption{1-norm and Frobenius norm (upper) of the difference between the true $\Psi^*$ and approximated $\widetilde{\Psi}$, and their computation time (lower), for various $T$ and $S$.}
    \vspace{0.3cm}
    \label{tab:approx_psi}
    \begin{tabular}{lccc}
    \hline \text { $(T,S)$} & \text { $(5, 5)$ } &
    \text { $(10, 10)$ } & \text { $(50, 10)$ } \\
    \hline
    \hline
    \text { $\| \Psi^* - \widetilde{\Psi}\|_1$ } & 0.118 & 0.039 & 0.022 \\
    \text { $\| \Psi^* - \widetilde{\Psi}\|_F$ } & 0.162 & 0.062 & 0.041 \\
    \hline
    \text { Time $\Psi^*$ (s.) } & 582& 4952 & 25343\\
    \text { Time $\widetilde{\Psi}$ (s.) } & 0.18 & 7.5 & 45.7 \\
    \hline
    \end{tabular}
\end{table*}
}}
The results validate the approximation choice for $\Psi$: as $T$ and $S$ grow, the norms tend to $0$ with a linear rate. As expected, the computational time for the true $\Psi$ explodes, while it remains feasible for the approximated version.

\section{Applications to seismic data}\label{sec:app}

Seismological data is made available by each country most impacted by seismic events. The Northern California
Earthquake Data Center (NCEDC, \cite{ncedc}) provides time series datasets, collecting information such as location or time about seismic events in California. Studying seismic regions reveals time series with highly complex dependence structures, that can be found between events and between neighboring regions \cite{Ogata1999, ogata1998space, vere1995forecasting}. Furthermore, predicting subsequent earthquakes is of greater importance. The first proposed method to study earthquake occurrences is the Epidemic Type Aftershock Sequence (ETAS; \cite{ogata1988statistical}) model relying only on the time values and their magnitude to model such data, ignoring the spatial dimension. Hawkes processes are well-suited to model earthquake occurrences \cite{musmeci1992space} due to their self-exciting nature in space and time: an earthquake can trigger further replica in a given period and spatial neighbourhood. These triggered events, often referenced as ‘aftershocks', can trigger other events. A space-time clustering form is generally observed when studying seismic datasets. Actual models usually assume space-time separated kernels with Gaussian density for the spatiality and an exponential density for the temporality, and thus limit the modelization power of such processes \cite{schoenberg2003multidimensional,veen2008estimation,zhuang2011next,fox2016spatially}.

We carry out experiments on three different datasets extracted from NCEDC database, each with a different clustering structure and behavior, and different lags of time. The first dataset \texttt{‘1968-1970'} contains $496$ events, the second dataset \texttt{‘1985-1986'} contains $998$ events, and the third dataset \texttt{‘1987-1989'} counts $605$ events, defined as seismic records with a magnitude larger than 3.0. Each event is defined by its time and location (no other information is used in our experiments).

Thanks to the flexibility of our proposed framework, we apply several kernels for both spatial and temporal components. We set $\Delta=(0.1, 0.1, 0.1)$, $W_{\mathcal{X}}=1$, $W_{\mathcal{Y}}=1$ and $W_{T}=1$ with truncated Gaussian (TG) and inverse Power-Law (POW, see Appendix \ref{sec:add:exp}) as spatial kernels and TG, exponential (EXP) and Kumaraswamy (KUM) as temporal ones. One can see that the best performance is achieved for different kernels on the different datasets, which confirms the limiting modelization power of the models used so far and the advantages of our approach.




{\renewcommand{\arraystretch}{1.5} 
{\setlength{\tabcolsep}{0.5cm}
\begin{table*}[!h]
    
    \caption{Negative Log-Likelihood values on test sets of various extracted datasets with several triggering kernels.}
    \label{tab:RD_1}
    \centering
    \begin{tabular}{l|ccc}
    \hline \text { Setting } & \texttt { 1968 - 1970 } & \texttt { 1985 - 1986 } & \texttt { 1987 - 1989 } \\
    \hline
    \hline
    \text { TG + TG } & 1.9530 & 2.5714 & 2.3367 \\
    \text { TG + EXP } & 1.0846 & 2.5116 & 2.8299 \\
    \text { TG + KUM } & {\bf 1.0771} & 2.4365 & 2.3627 \\
    \hline
    \text{ POW + TG} & 2.1350 & 1.8941 & 0.8227 \\
    \text{ POW + EXP} & 2.2257 & 1.3787 & 0.8215 \\
    \text{ POW + KUM} & 2.1657 & {\bf 1.3305} & {\bf 0.7840} \\
    \hline
    \end{tabular}
\end{table*}
}}

\section{Conclusion and perspectives}\label{sec:conclusion}

Motivated by the growing demand for spatio-temporal data modeling in various fields, we have introduced a novel, flexible and efficient approach to infer any parametric kernels in the context of space-time Hawkes processes that are not necessarily independent of space-time. Based on ERM-inspired least squares loss for point processes, the framework overcomes the significant computational difficulties inherent in fitting such complex models. It  relies on the use of finite support kernels, a discretization scheme and precomputations. After showing theoretically that the discretization error is minimal, both theoretically and numerically, we have investigated numerically the statistical efficiency of our approach. As the precomputation terms are expansive, we propose a computationally efficient approximation and show that the error is negligible. Finally, we demonstrate the value of using different kernels to model earthquake aftershocks, which is possible thanks to the numerical/statistical advantages of our method.

\bibliography{biblio}

\begin{thebibliography}{40}
\providecommand{\natexlab}[1]{#1}
\providecommand{\url}[1]{\texttt{#1}}
\expandafter\ifx\csname urlstyle\endcsname\relax
  \providecommand{\doi}[1]{doi: #1}\else
  \providecommand{\doi}{doi: \begingroup \urlstyle{rm}\Url}\fi

\bibitem[nce(2014)]{ncedc}
Northern {C}alifornia {E}arthquake {D}ata {C}enter, {N}{C}{E}{D}{C} {D}ataset.
\newblock \emph{UC Berkeley Seismological Laboratory}, 2014.
\newblock \doi{doi:10.7932/NCEDC}.

\bibitem[Bacry et~al.(2020)Bacry, Bompaire, Ga{\"\i}ffas, and Muzy]{bacry2020sparse}
Bacry, E., Bompaire, M., Ga{\"\i}ffas, S., and Muzy, J.-F.
\newblock Sparse and low-rank multivariate {Hawkes} processes.
\newblock \emph{Journal of Machine Learning Research}, 21\penalty0 (50):\penalty0 1--32, 2020.

\bibitem[Bompaire(2019)]{bompaire:tel-02316143}
Bompaire, M.
\newblock \emph{{Machine learning based on Hawkes processes and stochastic optimization}}.
\newblock Theses, {Universit{\'e} Paris Saclay (COmUE)}, July 2019.
\newblock URL \url{https://tel.archives-ouvertes.fr/tel-02316143}.

\bibitem[Brandolini et~al.(2013)Brandolini, Colzani, Gigante, and Travaglini]{brandolini2013koksma}
Brandolini, L., Colzani, L., Gigante, G., and Travaglini, G.
\newblock On the koksma--hlawka inequality.
\newblock \emph{Journal of Complexity}, 29\penalty0 (2):\penalty0 158--172, 2013.

\bibitem[Choi \& Hall(1999)Choi and Hall]{choi1999nonparametric}
Choi, E. and Hall, P.
\newblock Nonparametric approach to analysis of space-time data on earthquake occurrences.
\newblock \emph{Journal of Computational and Graphical Statistics}, 8\penalty0 (4):\penalty0 733--748, 1999.

\bibitem[Daley et~al.(2003)Daley, Vere-Jones, et~al.]{daley2003introduction}
Daley, D.~J., Vere-Jones, D., et~al.
\newblock \emph{An introduction to the theory of point processes: volume I: elementary theory and methods}.
\newblock Springer, 2003.

\bibitem[Diggle et~al.(1995)Diggle, Chetwynd, H{\"a}ggkvist, and Morris]{diggle1995second}
Diggle, P.~J., Chetwynd, A.~G., H{\"a}ggkvist, R., and Morris, S.~E.
\newblock Second-order analysis of space-time clustering.
\newblock \emph{Statistical methods in medical research}, 4\penalty0 (2):\penalty0 124--136, 1995.

\bibitem[Dong et~al.(2023)Dong, Zhu, Xie, Mateu, and Rodr{\'\i}guez-Cort{\'e}s]{dong2023non}
Dong, Z., Zhu, S., Xie, Y., Mateu, J., and Rodr{\'\i}guez-Cort{\'e}s, F.~J.
\newblock Non-stationary spatio-temporal point process modeling for high-resolution covid-19 data.
\newblock \emph{Journal of the Royal Statistical Society Series C: Applied Statistics}, 72\penalty0 (2):\penalty0 368--386, 2023.

\bibitem[D’Angelo et~al.(2022)D’Angelo, Payares, Adelfio, and Mateu]{dangelo}
D’Angelo, N., Payares, D., Adelfio, G., and Mateu, J.
\newblock Self-exciting point process modelling of crimes on linear networks.
\newblock \emph{Statistical Modelling}, 0\penalty0 (0):\penalty0 1471082X221094146, 2022.
\newblock \doi{10.1177/1471082X221094146}.
\newblock URL \url{https://doi.org/10.1177/1471082X221094146}.

\bibitem[Fox et~al.(2016)Fox, Schoenberg, and Gordon]{fox2016spatially}
Fox, E.~W., Schoenberg, F.~P., and Gordon, J.~S.
\newblock Spatially inhomogeneous background rate estimators and uncertainty quantification for nonparametric hawkes point process models of earthquake occurrences.
\newblock 2016.

\bibitem[Gonz{\'a}lez et~al.(2016)Gonz{\'a}lez, Rodr{\'\i}guez-Cort{\'e}s, Cronie, and Mateu]{gonzalez2016spatio}
Gonz{\'a}lez, J.~A., Rodr{\'\i}guez-Cort{\'e}s, F.~J., Cronie, O., and Mateu, J.
\newblock Spatio-temporal point process statistics: a review.
\newblock \emph{Spatial Statistics}, 18:\penalty0 505--544, 2016.

\bibitem[Hawkes(2018)]{Hawkes}
Hawkes, A.~G.
\newblock {Point Spectra of Some Mutually Exciting Point Processes}.
\newblock \emph{Journal of the Royal Statistical Society: Series B (Methodological)}, 33\penalty0 (3):\penalty0 438--443, 12 2018.
\newblock ISSN 0035-9246.
\newblock \doi{10.1111/j.2517-6161.1971.tb01530.x}.
\newblock URL \url{https://doi.org/10.1111/j.2517-6161.1971.tb01530.x}.

\bibitem[Holbrook et~al.(2022)Holbrook, Ji, and Suchard]{holbrook2022viral}
Holbrook, A.~J., Ji, X., and Suchard, M.~A.
\newblock From viral evolution to spatial contagion: a biologically modulated hawkes model.
\newblock \emph{Bioinformatics}, 38\penalty0 (7):\penalty0 1846--1856, 2022.

\bibitem[Ilhan \& Kozat(2020)Ilhan and Kozat]{ilhan2020modeling}
Ilhan, F. and Kozat, S.~S.
\newblock Modeling of spatio-temporal hawkes processes with randomized kernels.
\newblock \emph{IEEE Transactions on Signal Processing}, 68:\penalty0 4946--4958, 2020.

\bibitem[Kirchner(2017)]{kirchner2017estimation}
Kirchner, M.
\newblock An estimation procedure for the hawkes process.
\newblock \emph{Quantitative Finance}, 17\penalty0 (4):\penalty0 571--595, 2017.

\bibitem[Kresin et~al.(2022)Kresin, Schoenberg, and Mohler]{kresin2022comparison}
Kresin, C., Schoenberg, F.~P., and Mohler, G.
\newblock Comparison of hawkes and seir models for the spread of covid-19.
\newblock \emph{Advances and Applications in Statistics}, 74:\penalty0 83--106, 2022.

\bibitem[Kwon et~al.(2023)Kwon, Zheng, and Jun]{kwon2023flexible}
Kwon, J., Zheng, Y., and Jun, M.
\newblock Flexible spatio-temporal hawkes process models for earthquake occurrences.
\newblock \emph{Spatial Statistics}, 54:\penalty0 100728, 2023.

\bibitem[Laub et~al.(2015)Laub, Taimre, and Pollett]{laub2015hawkes}
Laub, P.~J., Taimre, T., and Pollett, P.~K.
\newblock Hawkes processes.
\newblock \emph{arXiv preprint arXiv:1507.02822}, 2015.

\bibitem[Lewis \& Mohler(2011)Lewis and Mohler]{lewis2011nonparametric}
Lewis, E. and Mohler, G.
\newblock A nonparametric em algorithm for multiscale hawkes processes.
\newblock \emph{Journal of nonparametric statistics}, 1\penalty0 (1):\penalty0 1--20, 2011.

\bibitem[Mohler(2014)]{mohler2014marked}
Mohler, G.
\newblock Marked point process hotspot maps for homicide and gun crime prediction in chicago.
\newblock \emph{International Journal of Forecasting}, 30\penalty0 (3):\penalty0 491--497, 2014.

\bibitem[Mohler et~al.(2011)Mohler, Short, Brantingham, Schoenberg, and Tita]{mohler2011self}
Mohler, G.~O., Short, M.~B., Brantingham, P.~J., Schoenberg, F.~P., and Tita, G.~E.
\newblock Self-exciting point process modeling of crime.
\newblock \emph{Journal of the American Statistical Association}, 106\penalty0 (493):\penalty0 100--108, 2011.

\bibitem[M{\o}ller \& Rasmussen(2005)M{\o}ller and Rasmussen]{moller2005perfect}
M{\o}ller, J. and Rasmussen, J.~G.
\newblock Perfect simulation of hawkes processes.
\newblock \emph{Advances in applied probability}, 37\penalty0 (3):\penalty0 629--646, 2005.

\bibitem[M\"uller \& Mateu(2012)M\"uller and Mateu]{book_spatial}
M\"uller, W. and Mateu, J.
\newblock \emph{Spatio-temporal Design: Advances in Efficient Data Acquisition}.
\newblock Wiley, 2012.

\bibitem[Musmeci \& Vere-Jones(1992)Musmeci and Vere-Jones]{musmeci1992space}
Musmeci, F. and Vere-Jones, D.
\newblock A space-time clustering model for historical earthquakes.
\newblock \emph{Annals of the Institute of Statistical Mathematics}, 44:\penalty0 1--11, 1992.

\bibitem[Ogata(1988)]{ogata1988statistical}
Ogata, Y.
\newblock Statistical models for earthquake occurrences and residual analysis for point processes.
\newblock \emph{Journal of the American Statistical association}, 83\penalty0 (401):\penalty0 9--27, 1988.

\bibitem[Ogata(1998)]{ogata1998space}
Ogata, Y.
\newblock Space-time point-process models for earthquake occurrences.
\newblock \emph{Annals of the Institute of Statistical Mathematics}, 50:\penalty0 379--402, 1998.

\bibitem[Ogata(1999)]{Ogata1999}
Ogata, Y.
\newblock \emph{Seismicity Analysis through Point-process Modeling: A Review}, pp.\  471--507.
\newblock Birkh{\"a}user Basel, Basel, 1999.
\newblock ISBN 978-3-0348-8677-2.
\newblock \doi{10.1007/978-3-0348-8677-2_14}.
\newblock URL \url{https://doi.org/10.1007/978-3-0348-8677-2_14}.

\bibitem[Rambhatla et~al.(2022)Rambhatla, Zeighami, Shahabi, Shahabi, and Liu]{rambhatla2022toward}
Rambhatla, S., Zeighami, S., Shahabi, K., Shahabi, C., and Liu, Y.
\newblock Toward accurate spatiotemporal covid-19 risk scores using high-resolution real-world mobility data.
\newblock \emph{ACM Transactions on Spatial Algorithms and Systems (TSAS)}, 8\penalty0 (2):\penalty0 1--30, 2022.

\bibitem[Reinhart(2018)]{reinhart2018review}
Reinhart, A.
\newblock A review of self-exciting spatio-temporal point processes and their applications.
\newblock \emph{Statistical Science}, 33\penalty0 (3):\penalty0 299--318, 2018.

\bibitem[Reynaud-Bouret \& Rivoirard(2010)Reynaud-Bouret and Rivoirard]{reynaud2010near}
Reynaud-Bouret, P. and Rivoirard, V.
\newblock Near optimal thresholding estimation of a poisson intensity on the real line.
\newblock \emph{Electronic journal of statistics}, 4:\penalty0 172--238, 2010.

\bibitem[{Reynaud-Bouret} et~al.(2014){Reynaud-Bouret}, Rivoirard, Grammont, and {Tuleau-Malot}]{Reynaud-Bouret2014}
{Reynaud-Bouret}, P., Rivoirard, V., Grammont, F., and {Tuleau-Malot}, C.
\newblock Goodness-of-fit tests and nonparametric adaptive estimation for spike train analysis.
\newblock \emph{The Journal of Mathematical Neuroscience}, 4\penalty0 (1):\penalty0 1--41, 2014.

\bibitem[Schoenberg(2003)]{schoenberg2003multidimensional}
Schoenberg, F.~P.
\newblock Multidimensional residual analysis of point process models for earthquake occurrences.
\newblock \emph{Journal of the American Statistical Association}, 98\penalty0 (464):\penalty0 789--795, 2003.

\bibitem[Sheen et~al.(2022)Sheen, Zhu, and Xie]{sheen2022tensor}
Sheen, H., Zhu, X., and Xie, Y.
\newblock Tensor kernel recovery for discrete spatio-temporal hawkes processes.
\newblock \emph{IEEE Transactions on Signal Processing}, 70:\penalty0 5859--5870, 2022.

\bibitem[Staerman et~al.(2023)Staerman, Allain, Gramfort, and Moreau]{staerman2023fadin}
Staerman, G., Allain, C., Gramfort, A., and Moreau, T.
\newblock Fadin: Fast discretized inference for hawkes processes with general parametric kernels.
\newblock In \emph{International Conference on Machine Learning}, pp.\  32575--32597. PMLR, 2023.

\bibitem[Veen \& Schoenberg(2008)Veen and Schoenberg]{veen2008estimation}
Veen, A. and Schoenberg, F.~P.
\newblock Estimation of space--time branching process models in seismology using an em--type algorithm.
\newblock \emph{Journal of the American Statistical Association}, 103\penalty0 (482):\penalty0 614--624, 2008.

\bibitem[Vere-Jones(1995)]{vere1995forecasting}
Vere-Jones, D.
\newblock Forecasting earthquakes and earthquake risk.
\newblock \emph{International Journal of Forecasting}, 11\penalty0 (4):\penalty0 503--538, 1995.

\bibitem[Yuan et~al.(2019)Yuan, Li, Bertozzi, Brantingham, and Porter]{yuan2019multivariate}
Yuan, B., Li, H., Bertozzi, A.~L., Brantingham, P.~J., and Porter, M.~A.
\newblock Multivariate spatiotemporal hawkes processes and network reconstruction.
\newblock \emph{SIAM Journal on Mathematics of Data Science}, 1\penalty0 (2):\penalty0 356--382, 2019.

\bibitem[Yuan et~al.(2021)Yuan, Schoenberg, and Bertozzi]{yuan2021fast}
Yuan, B., Schoenberg, F.~P., and Bertozzi, A.~L.
\newblock Fast estimation of multivariate spatiotemporal hawkes processes and network reconstruction.
\newblock \emph{Annals of the Institute of Statistical Mathematics}, pp.\  1--26, 2021.

\bibitem[Zhu \& Xie(2022)Zhu and Xie]{zhu2022spatiotemporal}
Zhu, S. and Xie, Y.
\newblock Spatiotemporal-textual point processes for crime linkage detection.
\newblock \emph{The Annals of Applied Statistics}, 16\penalty0 (2):\penalty0 1151--1170, 2022.

\bibitem[Zhuang(2011)]{zhuang2011next}
Zhuang, J.
\newblock Next-day earthquake forecasts for the japan region generated by the etas model.
\newblock \emph{Earth, planets and space}, 63:\penalty0 207--216, 2011.

\end{thebibliography}
\bibliographystyle{style}

\newpage

\appendix

\counterwithin{table}{section}
\counterwithin{figure}{section}

\section{Technical details}

\subsection{Gradients derivation}\label{sec:grad}

Here we derive the gradients of the proposed loss w.r.t. each set of parameters. Let $\tau=(\tau_x,\tau_y, \tau_t)$ be a vector on the grid $\mathcal{G}$.

\paragraph{Baselines.} The gradient of the loss with respect to the constant background for all $m \in \intervalleEntier{1}{D}$ is

{\small
\begin{align*}
    \frac{\partial \mathcal{L}_{\mathcal{G}}(\boldsymbol{\theta}, \widetilde{\mathcal{H}}_{T})}{\partial \boldsymbol{\mu}_m} & = 2 (T+\Delta_T) (2 S_{\mathcal{X}} + \Delta_{\mathcal{X}})(2 S_{\mathcal{Y}} + \Delta_{\mathcal{Y}})  \boldsymbol{\mu}_m  - 2 N_T^m  \\& + 2 \Delta_{\mathcal{X}}\Delta_{\mathcal{Y}} \Delta_T \sum\limits_{k=1}^D \sum\limits_{\tau_x = 1}^{L_{\mathcal{X}}} \sum\limits_{\tau_y = 1}^{L_{\mathcal{Y}}}  \sum\limits_{\tau_t=1}^{L_T} \boldsymbol{\alpha}_{mk} \, g_{mk}^{\Delta}\left[\tau\right] \Phi_k( \tau; G).
\end{align*}}

\paragraph{Excitation scaling parameters.}
The gradient of the loss with respect to $\boldsymbol{\alpha}_{m, l}$ for all $(m, l) \in \intervalleEntier{1}{D}^2$ is

{\small
\begin{align*}
    \frac{\partial \mathcal{L}_{\mathcal{G}}(\boldsymbol{\theta}, \widetilde{\mathcal{H}}_{T})}{\partial \boldsymbol{\alpha}_{m, l}} & = 2 \Delta_{\mathcal{X}}\Delta_{\mathcal{Y}} \Delta_T \boldsymbol{\mu}_m \sum\limits_{\tau_x = 1}^{L_{\mathcal{X}}} \sum\limits_{\tau_y = 1}^{L_{\mathcal{Y}}} \sum\limits_{\tau_t=1}^{L_T} g_{ml}^{\Delta}\left[\tau\right] \Phi_l(\tau; G)
    - 2 \sum\limits_{\tau_x = 1}^{L_{\mathcal{X}}} \sum\limits_{\tau_y = 1}^{L_{\mathcal{Y}}} \sum\limits_{\tau_t=1}^{L_T} g_{ml}^{\Delta}\left[\tau\right] \Phi_l( \tau; \widetilde{\mathcal{H}}_{ T}^m) \\&
    + 2 \Delta_{\mathcal{X}}\Delta_{\mathcal{Y}} \Delta_T \sum\limits_{k=1}^D \sum\limits_{\tau_x, \tau_x' = 1}^{L_{\mathcal{X}}} \sum\limits_{\tau_y, \tau_y' = 1}^{L_{\mathcal{Y}}} \sum\limits_{\tau_t, \tau_t'=1}^{L_T} \boldsymbol{\alpha}_{mk} \, g_{ml}^{\Delta}\left[\tau\right] g_{mk}^{\Delta}\left[\tau'\right] \Psi_{l, k}(\tau, \tau'; G).
\end{align*}}

\paragraph{Kernel parameters.}
The gradient of the loss with respect to the parameter of the kernel for all $(m, l) \in \intervalleEntier{1}{D}^2$ is

{\small
\begin{align*}
     \frac{\partial \mathcal{L}_{\mathcal{G}}(\boldsymbol{\theta}, \widetilde{\mathcal{H}}_{T})}{\partial \boldsymbol{\eta}_{m, l}} & = 2 \Delta_{\mathcal{X}} \Delta_{\mathcal{Y}}\Delta_T \boldsymbol{\mu}_m \sum\limits_{\tau_x = 1}^{L_{\mathcal{X}}} \sum\limits_{\tau_y = 1}^{L_{\mathcal{Y}}} \sum\limits_{\tau_t=1}^{L_T} \boldsymbol{\alpha}_{ml} \, \frac{\partial g_{ml}^{\Delta}\left[\tau\right]}{\partial \boldsymbol{\eta}_{m, l}} \Phi_l(\tau; G)
      \\&    - 2 \sum\limits_{\tau_x = 1}^{L_{\mathcal{X}}} \sum\limits_{\tau_y = 1}^{L_{\mathcal{Y}}}  \sum\limits_{\tau_t=1}^{L_T} \boldsymbol{\alpha}_{ml} \, \frac{\partial g_{ml}^{\Delta}\left[\tau\right]}{\partial \boldsymbol{\eta}_{m, l}} \Phi_l(\tau; \widetilde{\mathcal{H}}_{ T}^m) \\&
     + 2 \Delta_{\mathcal{X}} \Delta_{\mathcal{Y}}\Delta_T \sum\limits_{k=1}^D \sum\limits_{\tau_x, \tau_x' = 1}^{L_{\mathcal{S}}} \sum\limits_{\tau_y, \tau_y' = 1}^{L_{\mathcal{S}}} \sum\limits_{\tau_t, \tau_t'=1}^{L_T} \boldsymbol{\alpha}_{ml} \, \boldsymbol{\alpha}_{mk} \, \frac{\partial g_{ml}^{\Delta}\left[\tau\right]}{\partial \boldsymbol{\eta}_{m, l}} g_{mk}^{\Delta}\left[\tau'\right] \Psi_{l, k}(\tau, \tau'; G).
\end{align*}}

\subsection{Proof of Proposition \ref{prop:bias}}\label{subsec:prop:bias}

Denote by $\delta_{x, n}^i, \delta_{y, n}^i, \delta_{t, n}^i$ the spread between the original events and the $n$-th projected event of the $i$-th process, i.e.  $\tilde{x}_n^i   = x_n^i + \delta_{x, n}^i$, $\; \tilde{y}_n^i  = y_n^i + \delta_{y, n}^i$, $ \; \tilde{t}_n^i  = t_n^i +  \delta_{t, n}^i $ with $\delta_{x, n}^i\in [-\Delta_{\mathcal{X}}/2, \Delta_{\mathcal{X}}/2],  \delta_{y, n}^i  \in[-\Delta_{\mathcal{Y}}/2, \Delta_{\mathcal{Y}}/2] $ and $\delta_{t, n}^i\in [-\Delta_T/2, \Delta_T/2]$. Recall that for any $v=(v_x, v_y, v_t)\in \intervalleEntier{0}{G_{\mathcal{X}}} \times \intervalleEntier{0}{G_{\mathcal{Y}}} \times \intervalleEntier{0}{G_T}$, we have
\begin{align}
    \lambda_i(v_x\Delta_{\mathcal{X}}, v_y\Delta_{\mathcal{Y}}, v_t\Delta_T) &= \mu_i + \sum_{j=1}^p \sum_{u_m^j\in \mathcal{H}_T^j}g_{ij}(v_x\Delta_{\mathcal{X}}- x_m^j, v_y\Delta_{\mathcal{Y}}- y_m^j, v_t\Delta_T- t_m^j) . \nonumber
\end{align}

By defining $\Delta=(\Delta_{\mathcal{X}}, \Delta_{\mathcal{Y}}, \Delta_T)$, $\delta^i_n=(\delta_{x, n}^i, \delta_{y, n}^i, \delta_{t, n}^i), \; \forall ~i\in \llbracket 1, D \rrbracket$ and $\forall ~1\leq n\leq N_T^i$, the vector of the intensity function on the grid is given by
\begin{align}
    \tilde{\lambda}_i[v] &= \mu_i + \sum_{j=1}^{D} \sum_{\tilde{u}_m^j\in  \tilde{\mathcal{H}}_{v\Delta}^j} g_{ij} (v\Delta - \tilde{u}_m^j)\nonumber \\
    &= \mu_i + \sum_{j=1}^{D} \sum_{u_m^j\in  \mathcal{H}_{v\Delta}^j} g_{ij} (v\Delta - u_m^j - \delta_m^j) \label{bias:l3}
\end{align}

where \ref{bias:l3} holds because $\Delta_{\mathcal{X}} < \min\limits_{x_n^i, x_m^j \in \mathcal{H}_T} |x_n^i -x_m^j|$, $\Delta_{\mathcal{Y}} < \min\limits_{y_n^i, y_m^j \in \mathcal{H}_T} |y_n^i -y_m^j|$, and $\Delta_T < \min\limits_{t_n^i, t_m^j \in \mathcal{H}_T} |t_n^i -t_m^j|$, which ensures that no event collapses on the same bin of the grid and that $|\widetilde{\mathcal H}_{v\Delta}^j| = |\mathcal H_{v\Delta}^j|$, where $| \cdot|$ denotes the cardinal of a set. Note that this hypothesis also implies that the intensity function is smooth for all points on the grid $\mathcal G$. Applying the first-order Taylor expansion to the kernels $g_{ij}$ in $v\Delta - u_m^j$ and bounding the perturbation $\delta_m^j$ by $\Delta$ yields the first result of the proposition.

For the perturbation of the discrete loss, we have

\begin{align*}
    \mathcal{L}_{\mathcal{G}}(\boldsymbol{\theta}, \widetilde{\mathcal{H}}_{T}) &=  \sum\limits_{i=1}^D \left(\Delta_{\mathcal{X}}\Delta_{\mathcal{Y}} \Delta_T \sum\limits_{v_x = 0}^{G_{\mathcal{X}}} \sum\limits_{v_y = 0}^{G_{\mathcal{Y}}}\sum\limits_{v_t = 0}^{G_T} \left(\Tilde{\lambda}_i[v]\right)^2 - 2 \sum\limits_{\Tilde{u}_n^i \in \widetilde{\mathcal{H}}_{ T}^i} \Tilde{\lambda}_i\left[\frac{\Tilde{u}_n^i}{\Delta}\right]\right) \\&
     = \mathcal L(\theta, \mathcal{H}_{T}) + \sum_{i=1}^{D}\Bigg( \underbrace{\Delta_{\mathcal{X}}\Delta_{\mathcal{Y}} \Delta_T \sum\limits_{v_x= 0}^{G_{\mathcal{X}}}\sum\limits_{v_y= 0}^{G_{\mathcal{Y}}} \sum\limits_{v_t = 0}^{G_T}\tilde{\lambda}_{i}[v]^2 - \int_0^T  \int_{\mathcal{S}} \lambda_i(x, y, t)^2 \mathrm{d}x\mathrm{d}y\mathrm{d}t}_{(*)}
        \\& \qquad \qquad \qquad \qquad \quad - 2\underbrace{\sum_{u_n^{i} \in \mathcal{H}_T^i} \bigg (\widetilde{\lambda}_{i}\Bigg[\frac{\widetilde u_n^i}{\Delta}\Bigg] - \lambda_{i}\pars{u_n^{i}}}_{(**)} \bigg)\Bigg),
\end{align*}

where $\frac{\Tilde{u}_n^i}{\Delta}$ is the division term by term of these three dimensional vectors. The first term $(*)$ is the error of a Riemann approximation of the integral. We can use the generalization of the Koksma-Hlawka inequality \cite{brandolini2013koksma} for piece-wise smooth functions on  compact set of $\mathbb{R}^d$:

\begin{equation}\label{term1}
   \Bigg | \Delta_{\mathcal{X}} \Delta_{\mathcal{Y}} \Delta_T \sum\limits_{v_x= 0}^{G_{\mathcal{X}}} \sum\limits_{v_y= 0}^{G_{\mathcal{Y}}} \sum\limits_{v_t = 0}^{G_T} \tilde{\lambda}_{i}[v]^2 -\int_0^T \int_{\mathcal{S}}\lambda_i(x, y, t)^2 \mathrm{d}x\mathrm{d}y\mathrm{d}t \Bigg |\leq  C(\lambda_i) \|\Delta\|_2,
\end{equation}
where $\|\Delta\|_2$ comes from the maximal distance on the uniform spatio-temporal grid and $C(\lambda_i)$ is a constant that depends on the regularity of $\lambda_i$, see Theorem 1 in \cite{brandolini2013koksma}.

For the second term $(**)$, we re-use the expression from\ref{bias:l3} but use a Taylor expansion in $u_n^i - u_m^j$.
The perturbation becomes $\delta^j_m - \delta^i_n$,
\begin{equation}\label{term2}
    \sum_{u_n^{i} \in \mathcal{H}_T^i} \bigg( \widetilde{\lambda}_{i}\Bigg[\frac{\widetilde u_n^i}{\Delta}\Bigg] - \lambda_{i}\pars{u_n^{i}} \bigg)
    = \sum_{j=1}^D\sum_{ \substack{u_n^{i} \in \mathcal{H}_T^i \\ u_m^{j} \in \mathcal{H}_T^j}}\pars{\delta_n^i - \delta_m^j}
    \nabla_u g_{ij}\pars{u_n^i - u_m^j} + O\pars{\|\Delta\|_2^2}   .
\end{equation}
Summing \ref{term1} and \ref{term2} concludes the proof.

\subsection{Proof of Proposition \ref{prop2}}\label{subsec:prop2}

    We consider the two estimators $\widehat{\boldsymbol{\theta}}_\Delta = \argmin_{\boldsymbol{\theta}}~ \mathcal{L}_{\mathcal{G}}(\boldsymbol{\theta}, \widetilde{\mathcal{H}}_T)$ and $\widehat{\boldsymbol{\theta}}_c = \argmin_{\boldsymbol{\theta}}~\mathcal{L}(\boldsymbol{\theta}, \mathcal{H}_T)$.
    With the loss approximation from \ref{prop:bias}, we have a point-wise convergence of $\mathcal{L}_{\mathcal{G}}(\boldsymbol{\theta}, \widetilde{\mathcal{H}}_T)$ towards $\mathcal{L}(\boldsymbol{\theta}, \mathcal{H}_T)$ for all $\boldsymbol{\theta} \in \boldsymbol{\Theta}$ as $\|\Delta\|$ goes to 0.
    By continuity of $\mathcal{L}_{\mathcal{G}}(\boldsymbol{\theta}, \widetilde{\mathcal{H}}_T)$, we have that the limit of $\widehat{\boldsymbol{\theta}}_\Delta$ when $\|\Delta\|$ goes to 0 exists and is equal to $\widehat{\boldsymbol{\theta}}_c$.
    This proves that the discretized estimator converges to the continuous one as $\|\Delta\|$ decreases.

 The Karush-Kuhn-Tucker conditions imply that:
    \begin{equation}
        \label{eq:app:kkt}
        \nabla_{\boldsymbol{\theta}} \mathcal{L}_{\mathcal{G}}\pars{\widehat{\boldsymbol{\theta}}_\Delta, \widetilde{\mathcal{H}}_T} = 0 \qquad\text{and}\qquad \nabla_{\boldsymbol{\theta}}\mathcal{L}\pars{\widehat{\boldsymbol{\theta}}_c, \mathcal{H}_T} = 0 .
    \end{equation}

    Using the approximation from \ref{term1} and \ref{term2}, one gets in the limit of small $\|\Delta\|$:

\begin{align*}
   \nabla_{\boldsymbol{\theta}} \mathcal{L}_{\mathcal{G}}(\widehat{\boldsymbol{\theta}}_\Delta, \widetilde{\mathcal{H}}_{T}) \geq & \nabla_{\boldsymbol{\theta}} \mathcal{L}(\widehat{\boldsymbol{\theta}}_\Delta, \mathcal{H}_{T}) -\|\Delta\|_2 \sum_{i} \nabla_{\boldsymbol{\theta}}C(\lambda_i) \\& + 2 \sum_{i,j}\sum_{ \substack{u_n^{i} \in \mathcal{H}_T^i \\ u_m^{j} \in \mathcal{H}_T^j} }\pars{\delta_m^j - \delta_n^i}
    . \nabla_{\boldsymbol{\theta}} \nabla_{u} g_{ij}\pars{u_n^i - u_m^j} + O\pars{\|\Delta\|_2^2}.
\end{align*}

  Combining this with \ref{eq:app:kkt}, we get:

\begin{align*}
 \nabla_{\boldsymbol{\theta}} \mathcal{L} (\widehat{\boldsymbol{\theta}}_\Delta, \mathcal{H}_{T}) \leq &\|\Delta\|_2 \sum_{i}  \nabla_{\boldsymbol{\theta}}C(\lambda_i) \\& + 2 \sum_{i,j}\sum_{ \substack{u_n^{i} \in \mathcal{H}_T^i \\ u_m^{j} \in \mathcal{H}_T^j} }\pars{\delta_n^i - \delta_m^j}
    . \nabla_{\boldsymbol{\theta}} \nabla_{u} g_{ij}\pars{u_n^i - u_m^j} + O\pars{\|\Delta\|_2^2}.
\end{align*}

Thus, we have

\begin{align*}
    \norme{\nabla_{\boldsymbol{\theta}} \mathcal{L}(\widehat{\boldsymbol{\theta}}_\Delta, \mathcal{H}_{T}) - \nabla_{\boldsymbol{\theta}}\mathcal{L}\pars{\widehat{\boldsymbol{\theta}}_c, \mathcal{H}_T} }_2 & \leq \bigg | \bigg |  2 \sum_{i,j}\sum_{ \substack{u_n^{i} \in \mathcal{H}_T^i \\ u_m^{j} \in \mathcal{H}_T^j} }\pars{\delta_n^i - \delta_m^j}
    . \nabla_{\boldsymbol{\theta}} \nabla_{u} g_{ij}\pars{u_n^i - u_m^j} \\& \qquad +  \|\Delta\|_2\sum_{i}\nabla_{\boldsymbol{\theta}}C(\lambda_i)  \bigg | \bigg |_2 + O\pars{\|\Delta\|_2^2} \\&
    \leq \text{max}\{ \|\Delta\|_2, \|\Delta\|_{\infty}  \} ~\omega(\widehat{\boldsymbol{\theta}}_\Delta),
\end{align*}

where $\omega(\boldsymbol{\theta})= \bigg | \bigg |  2 \sum_{i,j}\sum_{ \substack{u_n^{i} \in \mathcal{H}_T^i \\ u_m^{j} \in \mathcal{H}_T^j} } \langle \mathbf{1}
    , \nabla_{\boldsymbol{\theta}} \nabla_{u} g_{ij}\pars{u_n^i - u_m^j} \rangle +  \sum_{i}\nabla_{\boldsymbol{\theta}}C(\lambda_i)  \bigg | \bigg |_2$ with $\mathbf{1}$ a three dimensional vector of one.   This function is a $O(1)$. Using the hypothesis that the hessian $\nabla^2_{\boldsymbol{\theta}}\mathcal L(\widehat{\boldsymbol{\theta}}_c, \mathcal{H}_T)$ exists and is positive definite with smallest eigenvalue $\varepsilon$, we have:
\begin{align*}
\varepsilon \norme{\widehat{\boldsymbol{\theta}}_\Delta - \widehat{\boldsymbol{\theta}}_c}_2^2
    &\le\norme{ \nabla_{\boldsymbol{\theta}} \mathcal L\pars{\widehat{\boldsymbol{\theta}}_\Delta, \mathcal{H}_{T}} - \nabla_{\boldsymbol{\theta}} \mathcal L\pars{\widehat{\boldsymbol{\theta}}_c, \mathcal{H}_{T}}}_2^2 \\
\text{\ie}\qquad
\varepsilon \norme{\widehat{\boldsymbol{\theta}}_\Delta - \widehat{\boldsymbol{\theta}}_c}_2^2
    &\le \dfrac{\mathrm{max}\{  \|\Delta\|_2, \|\Delta\|_{\infty} \}}{\varepsilon}~\omega\pars{\widehat{\boldsymbol{\theta}}_\Delta}.
\end{align*}
This concludes the proof.

\section{Additional experiments} \label{sec:add:exp}

This section presents complementary results related to  \ref{subsec:num_exp_dis}. We reproduce the same experiments with one additional kernel for the spatial triggering function, two additional kernels for the temporal triggering function, and give some details about each parameter estimation separately.

\begin{figure}
\centering
\begin{tabular}{cc}
    \multicolumn{2}{c}{\hspace{0.2cm} \includegraphics[scale=0.3]{images/legend/legend_exp1.png}}\\
    \includegraphics[scale = 0.4]{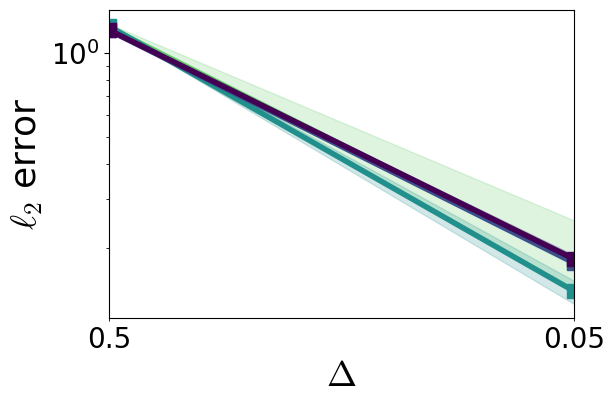}
           \end{tabular}
    \caption{Median and $25$\%-$75$\% quantiles of the $\ell_2$-norm between true and estimated parameters for a truncated Inverse Power Law spatial kernel and a Kumaraswamy temporal kernel, with respect to $\Delta$, for various $T$ and $S$.}
    \label{fig:exp1_POW_KUR}
\end{figure}

\paragraph{Additional spatial kernel.} To illustrate the flexibility of our proposed method, we show the obtained results with a new spatial triggering function: the truncated Inverse Power Law kernel, defined by
$$h(x, y \,; m_1,m_2, d) = \left(1 + \frac{(x-m_1)^2 + (y - m_2)^2}{d}\right)^{-3/2} \mathbb{I} \{(x, y) \in [-1, 1]^2\},$$
where the set of parameters to estimate is $\boldsymbol{\theta^*} = (\mu, \alpha, m, d, a, b)$. The temporal kernel function is chosen to be the Kumaraswamy function (defined in \ref{subsec:num_exp_dis}). In \ref{fig:exp1_POW_KUR}, we show the median (over $10$ runs) and the $25$\%-$75$\% quantiles of the $\ell_2$ estimation error, for various values of $T$, $S$ and $\Delta$ (with the same values as in \ref{subsec:num_exp_dis}).

One can observe that the $\ell_2$ error tends toward zero as $\Delta$ decreases, as for the Gaussian spatial kernel result in \ref{fig:exp1_kur}.

\paragraph{Additional temporal kernels.}  Now, we propose two new kernels for the temporal triggering function. We first investigate the popular case of the truncated Exponential kernel, with decay $\lambda = 1$, defined by
$$f(t \,; \lambda) \propto \lambda \exp \left(- \lambda t \right) \mathbb{I} \{0 \leq t \leq W_T \}.$$
Then, in the second setting, the kernel is the truncated Gaussian kernel (defined in \ref{subsec:num_exp_comp}). The spatial kernel is the truncated Gaussian kernel for both cases. Hence, the parameters to estimate are $\boldsymbol{\theta^*} = (\mu, \alpha, m, \sigma, \lambda)$ (resp. $\boldsymbol{\theta^*} = (\mu, \alpha, m, \sigma, m_T, \sigma_T)$) for the case with the truncated Exponential temporal kernel (resp. the truncated Gaussian kernel).

\begin{figure}[!ht]
    \centering
    \begin{tabular}{cc}
    \multicolumn{2}{c}{\hspace{0.6cm} \includegraphics[scale=0.3]{images/legend/legend_exp1.png}} \\
         \includegraphics[scale=0.4, trim=0cm 0 0 0]{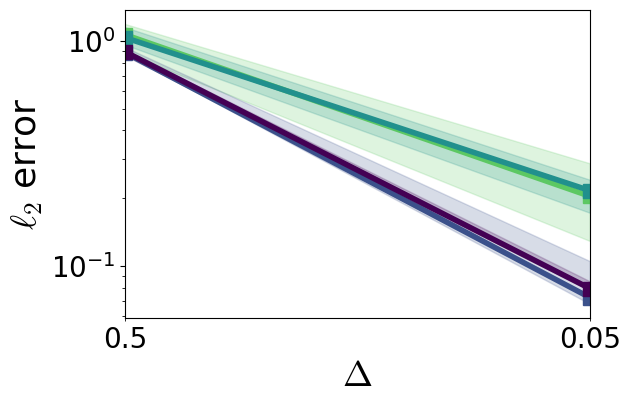} & \includegraphics[scale=0.4, trim=0cm 0 0 0]{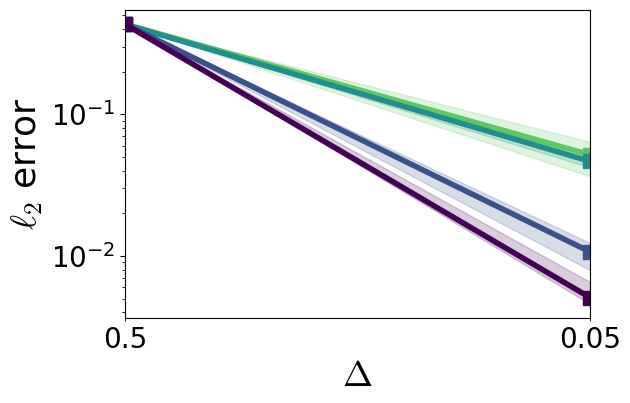}
           \end{tabular}
    \caption{Median and $25$\%-$75$\% quantiles of the $\ell_2$-norm between true and estimated parameters for a truncated Exponential (left), and a truncated Gaussian (right) temporal kernels, with respect to $\Delta$, for various $T$ and $S$.}
    \label{fig:exp1_Exp_Gauss}
\end{figure}

We apply the same experiments as in \ref{subsec:num_exp_dis}, and display the results in  \ref{fig:exp1_Exp_Gauss}: the median (over $100$ runs) and the $25\%-75\%$ quantiles of the $\ell_2$ estimation error $\| \hat{\boldsymbol{\theta}}- \boldsymbol{\theta}^* \|_2$ are given for the truncated Exponential kernel (left) and for the truncated Gaussian kernel (right).

We observe that, for both settings, the error decreases as the stepsizes increase. Furthermore, the error also decreases with respect to the values of $T$ and $S$.

We notice that the $\ell_2$ error is smaller in the case of a truncated Gaussian temporal kernel. This remark and the observations made for Figure \ref{fig:exp1_POW_KUR} support our claims: the method is efficient and flexible. Thus, this model can be well-suited for applications to real-world data, where the events do not immediately trigger more events and where the triggering structure does not necessarily follow a Gaussian function for the spatial domain.

\begin{figure}[!ht]
    \centering
    \begin{tabular}{cc}
    \multicolumn{2}{c}{\hspace{0.6cm} \includegraphics[scale=0.4]{images/legend/legend_exp1.png}} \\
    \hspace{0.8cm}$(\hat{\mu}-\mu)^2$ & \hspace{0.8cm}$(\hat{\alpha}-\alpha)^2$\\
         \includegraphics[scale=0.4, trim=0cm 0 0 0]{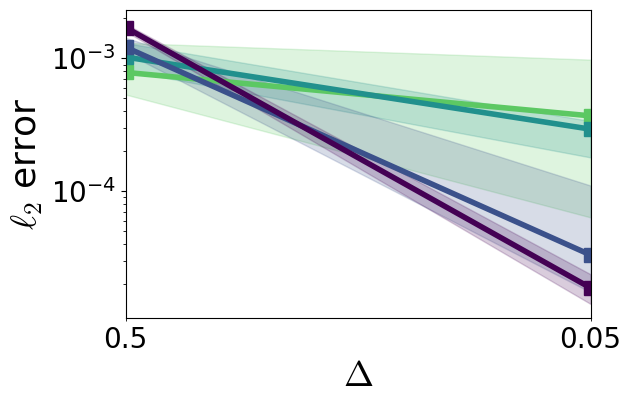} & \includegraphics[scale=0.4, trim=0cm 0 0 0]{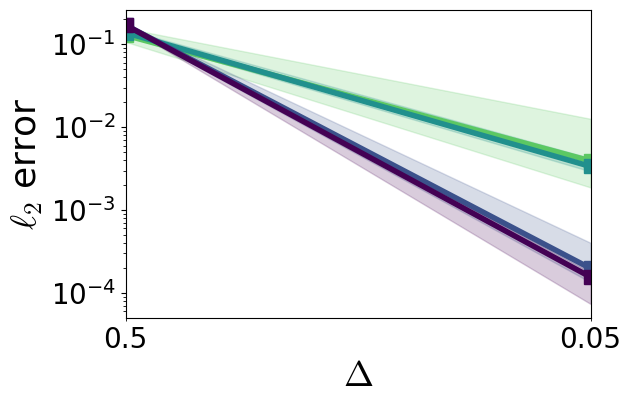} \\
         \hspace{0.8cm}$(\hat{m}-m)^2$ & \hspace{0.8cm}$(\hat{\sigma}-\sigma)^2$ \\
         \includegraphics[scale=0.4, trim=0cm 0 0 0]{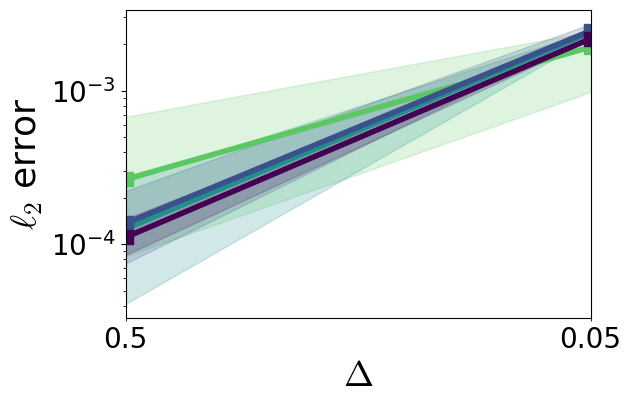} & \includegraphics[scale=0.4, trim=0cm 0 0 0]{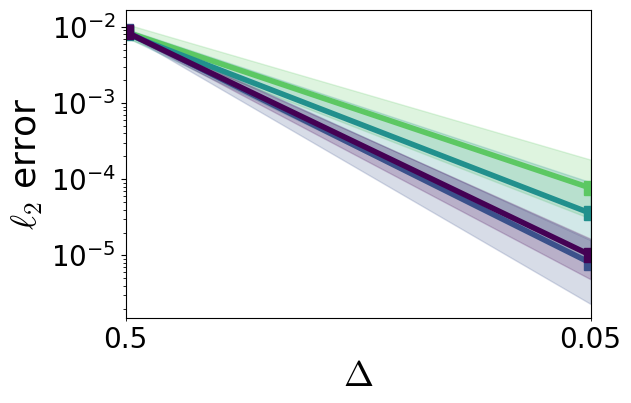}\\
         \hspace{0.8cm}$(\hat{a}-a)^2$ & \hspace{0.8cm}$(\hat{b}-b)^2$ \\
         \includegraphics[scale=0.4, trim=0cm 0 0 0]{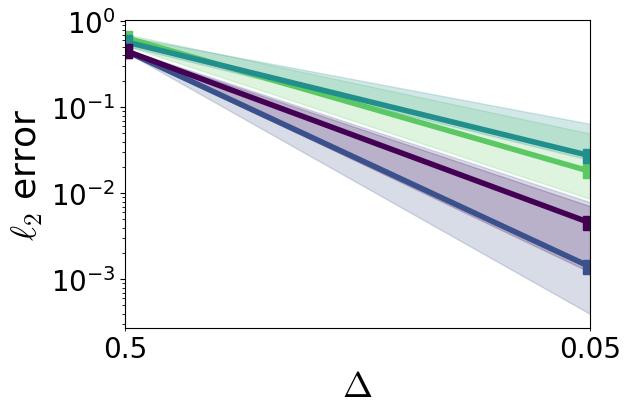} & \includegraphics[scale=0.4, trim=0cm 0 0 0]{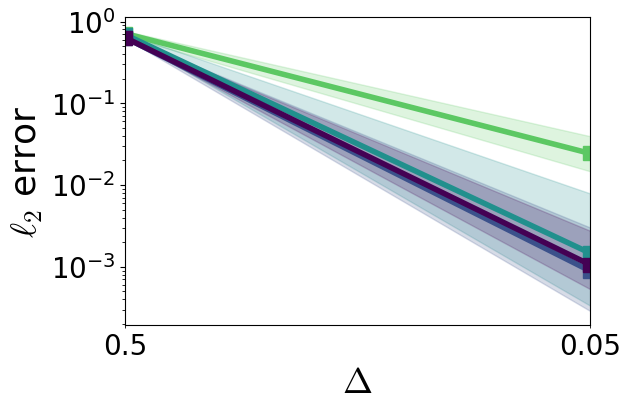}
           \end{tabular}
    \caption{Square error on parameters for the Kumaraswamy temporal kernel, as a function of $T$, $S$ and $\Delta$.}
    \label{fig:exp1_details_kur}
\end{figure}

\begin{figure}[!ht]
    \centering
    \begin{tabular}{cc}
    \multicolumn{2}{c}{\hspace{0.6cm} \includegraphics[scale=0.4]{images/legend/legend_exp1.png}} \\
    \hspace{0.8cm}$(\hat{\mu}-\mu)^2$ & \hspace{0.8cm}$(\hat{\alpha}-\alpha)^2$\\
        \includegraphics[scale=0.4, trim=0cm 0 0 0]{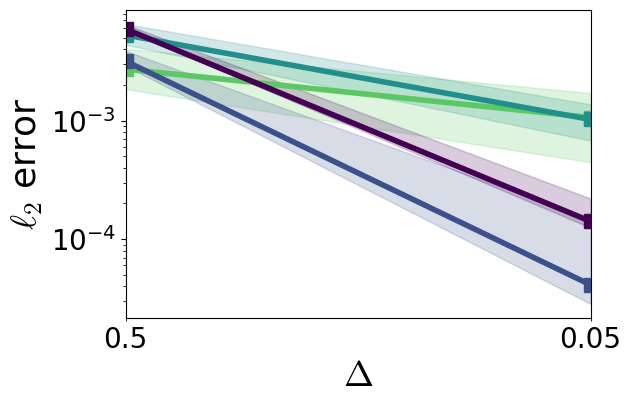}
    &
    \includegraphics[scale=0.4, trim=0cm 0 0 0]{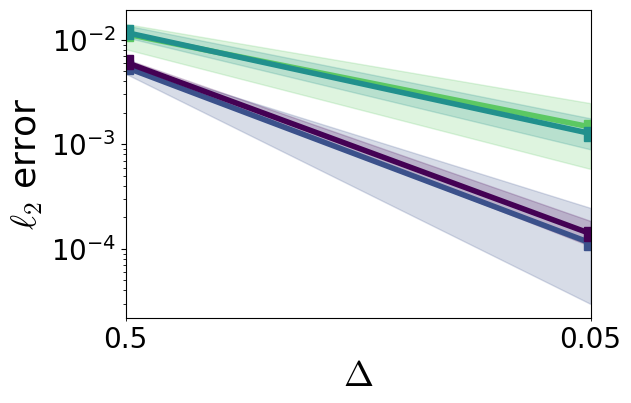}
    \\
         \hspace{0.8cm}$(\hat{m}-m)^2$ & \hspace{0.8cm}$(\hat{\sigma}-\sigma)^2$ \\
    \includegraphics[scale=0.4, trim=0cm 0 0 0]{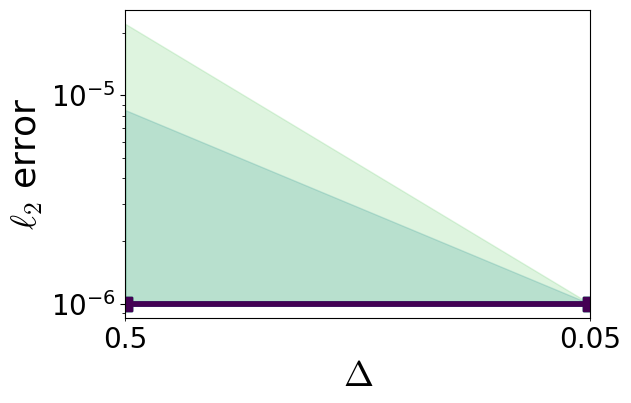}
    &
    \includegraphics[scale=0.4, trim=0cm 0 0 0]{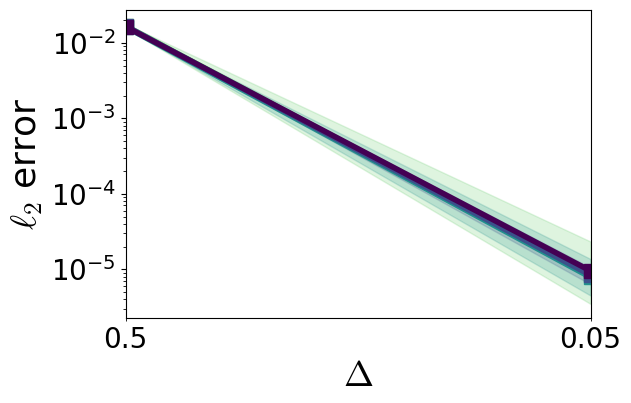}
    \\
         &\hspace{-7cm}$(\hat{\lambda}-\lambda)^2$  \\
    \includegraphics[scale=0.4, trim=0cm 0 15cm 0]{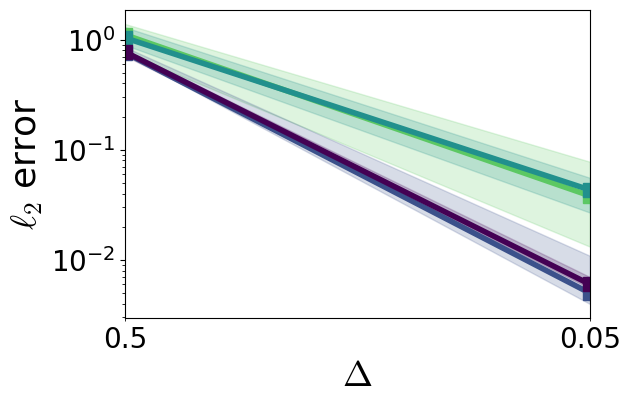}
    \end{tabular}
    \caption{Square error on parameters for the truncated Exponential temporal kernel, as a function of $T$, $S$ and $\Delta$.}
    \label{fig:exp1_details_exp}
\end{figure}

\paragraph{Details about parameter estimation.} In addition, we display the $\ell_2$ error for each parameter separately in Figure \ref{fig:exp1_details_kur} for the Kumaraswamy and in Figure \ref{fig:exp1_details_exp} for the truncated Exponential kernels.

\clearpage

\end{document}